\documentclass[12pt]{article}

\usepackage{amsmath}
\usepackage{numprint}
\usepackage{algorithm}
\usepackage{algpseudocode}
\algrenewcommand\algorithmicrequire{\textbf{Input:}}
\algrenewcommand\algorithmicensure{\textbf{Output:}}
\usepackage{amsmath,amssymb}

\usepackage[dvips]{graphicx,color}
\usepackage{natbib}
\usepackage{mathrsfs}
\usepackage{bm, xcolor}
\usepackage{mathrsfs}
\usepackage{verbatim}
\usepackage{longtable}
\usepackage{threeparttable}
\usepackage{color}
\usepackage{amsmath}

\usepackage{enumerate}
\usepackage{amssymb}
\usepackage{amsbsy}
\usepackage{amsmath}
\usepackage{amsthm}
\usepackage{amsfonts}
\usepackage{latexsym}
\usepackage{graphicx}
\usepackage{xifthen} % Allows us to put in optional arguments to questions
\usepackage{color}
\usepackage{manfnt}
\usepackage{ifthen}
\usepackage{bm}
\usepackage{caption}
\usepackage{subcaption}

\makeatletter
\def\singlespace{\def\baselinestretch{1}\@normalsize}

\newtheorem{lemma}{Lemma}%[section]
\newtheorem{theorem}{Theorem}%[section]

\newtheorem{example}{Example}%[section]
\@addtoreset{equation}{section}

\renewcommand{\hat}{\widehat}

\def\singlespace{\def\baselinestretch{1}\@normalsize}

\makeatother

\def\newpage{\vfill\eject}

\newdimen\biblioindent    \biblioindent=30pt

 at 7truept

\def\beqr{\begin{eqnarray}}
	\def\eeqr{\end{eqnarray}}
\def\beqrs{\begin{eqnarray*}}
	\def\eeqrs{\end{eqnarray*}}

\def\beq{\begin{equation}}
\def\eeq{\end{equation}}
\def\beqn{\begin{eqnarray}}
\def\eeqn{\end{eqnarray}}
\def\beqnn{\begin{eqnarray*}}
\def\eeqnn{\end{eqnarray*}}

\newtheorem*{assumption*}{\assumptionnumber}
\providecommand{\assumptionnumber}{}


\theoremstyle{definition}
\newtheorem{defi}{Definition}%[section]
\newtheorem{claim}{Claim}%[section]

\makeatletter

\newcommand{\Rmnum}[1]{\expandafter\@slowromancap\romannumeral #1@}
\makeatother
\begin{document}
	\renewcommand{\baselinestretch}{1.3}
	\newcommand{\xintao}[1]{
		{\color{blue!90} Xintao:  #1}}
	
	\title {\bf   Adaptive False Discovery Rate Control with Privacy Guarantee}
	\author{     Xintao Xia, Zhanrui Cai\\
		Department of Statistics, Iowa State University     }
	\date{\empty}
	
	\maketitle
	\renewcommand{\baselinestretch}{1.5}
	\baselineskip=24pt
	\noindent{\bf Abstract:}
	
Differentially private multiple testing procedures can protect the information of individuals used in hypothesis tests while guaranteeing a small fraction of false discoveries. In this paper, we propose a differentially private adaptive FDR control method that can control the classic FDR metric exactly at a user-specified level $\alpha$ with privacy guarantee, which is a non-trivial improvement compared to the differentially private Benjamini-Hochberg method proposed in \cite{dwork2021differentially}. Our analysis is based on two key insights: 1) a novel $p$-value transformation that preserves both privacy and the mirror conservative property, and 2) a mirror peeling algorithm that allows the construction of the filtration and application of the optimal stopping technique. Numerical studies demonstrate that the proposed DP-AdaPT performs better compared to the existing differentially private FDR control methods. Compared to the non-private AdaPT, it incurs a small accuracy loss but significantly reduces the computation cost.

	%Differential privacy data analysis has received growing attention in various applications, as well as modifying classic statistical methods with privacy guarantees. This paper proposes the first adaptive false discovery rate (FDR) control procedure with differential privacy constraints. The differential Privacy procedure begins with adding noise and selecting a subset of potential $p$-values for further analysis. The noisy selection procedure violates some necessary conditions in the classic FDR control procedure and thus creates difficulties in constructing new procedures. Motivated by the martingale-based adaptive procedure, we construct an ingenious adding noisy procedure and a novel differential private selection procedure. An adaptive and flexible procedure is performed after the selection. Our proposed procedure has a valid finite sample FDR control.
	
	\par \vspace{9pt} \noindent {\it Key words and phrases: selective inference; differential privacy; false discovery rate.}

	\newpage

	\renewcommand{\baselinestretch}{1.75}

\section{Introduction}

	\subsection{Differential privacy}
	
	With the advancement of technology, researchers are able to collect and analyze data on a large scale and make decisions based on data-driven techniques.	However, privacy issues could be encountered without a proper mechanism for data analysis and may lead to serious implications. For example, in bioinformatics, genomic data are usually very sensitive and irreplaceable, and it is of great importance to protect the individual's privacy in genome analysis, including GWAS \citep{kim2020privacy}. Many countries have classified genomic data as sensitive and must be handled according to certain regulations, such as the HIPAA in the USA and the Data Protection Directive in the European Union. The leak of individual genetic information can have severe consequences. It may underpin the trust of data-collecting agencies and discourage people or companies from sharing personal information. The leak of individual genetic information may also lead to serious social problems, such as genome-based discrimination \citep{kamm2013new}.
	
	In recent literature, a popular procedure to protect privacy is to apply {\it differentially private} algorithms in data analysis. First proposed by  \cite{dwork2006calibrating}, the concept of differential privacy has seen successful applications in numerous fields, including but not limited to healthcare, information management, government agencies, etc. Privacy is achieved by adding proper noise to the algorithm \citep{fuller1993masking} and obscuring each individual's characteristics. A differentially private procedure guarantees that an adversary can not judge whether a particular subject is included in the data set with high probability; thus is extremely useful in protecting personal information. During the past decades, considerable effort has been devoted to developing machine learning algorithms to guarantee differential privacy, such as differentially private deep learning \citep{abadi2016deep} or boosting \citep{dwork2010boosting}. In the statistics literature, \cite{wasserman2010statistical} estimated the convergence rate of density estimation in differential privacy. Other applications include but are not limited to differential privacy for functional data \citep{karwa2016inference}, network data \citep{karwa2016inference}, mean estimation, linear regression \citep{cai2021cost}, etc. More recently, \cite{dwork2021differentially} proposed the private Benjamini-Hochberg procedure to control the false discovery rate in multiple hypothesis testing. We refer readers to the classic textbook by \cite{dwork2014algorithmic} for a comprehensive review of differential privacy.

	\subsection{False Discovery Rate Control}
	
	In modern statistical analysis, large-scale tests are often conducted to answer research questions from scientists or the technology sector's management team. For example, in bioinformatics, researchers compare a phenotype to thousands of genetic variants and search for associations of potential biological interest. It is crucial to control the expected proportion of falsely rejected hypotheses, i.e., the  \textit{false discovery rate} \citep{benjamini1995controlling}. Controlling the false discovery rate (FDR) lets scientists increase power while maintaining a principled bound on the error. Let $R$ be the number of total rejections and $V$ be the number of false rejections; the FDR is defined as 
 \begin{equation}
    \label{eq: fdr}
	\mbox{FDR} = \mathbb{E}\left[ \frac{V}{\max\{R, 1\}}\right]. 
 \end{equation}
	
	The most famous multiple-testing procedure is the  Benjamini–Hochberg (BH) procedure \citep{benjamini1995controlling}. Given $n$ hypotheses and their ordered $p$-values $p_{(1)}< p_{(2)}< \dots< p_{(n)}$, the BH procedure rejects any null hypothesis whose $p$-value is non-greater than $\max\{p_{(i)}: p_{(i)}\leq \alpha i/n\}$, where $\alpha$ is a user-specified target level for FDR. \cite{benjamini2001control} extended the BH procedure to the setting where all the test statistics have positive regression dependency. Recent works focus on settings where prior information or extra data for hypotheses are available. The side information can be integrated to weighting the $p$-values \citep{genovese2006false, dobriban2015optimal}, exploring the group structures \citep{hu2010false} or natural ordering \citep{barber2015controlling, li2017accumulation} among hypotheses, etc. Adaptively focusing on the more promising hypotheses can also lead to a more powerful multiple-testing procedure \citep{lei2018adapt, tian2019addis}.
	
	Most multiple-testing procedures put assumptions on the $p$-values. A natural and mild assumption is that the $p$-values under the null hypothesis follow the uniform distribution on $[0,1]$. Because many statistical tests tend to be conservative under the null \citep{cai2022PNAS}, it is also common to assume that the $p$-values are stochastically larger than the uniform distribution, or {\it super-uniform}: $\mathbb{P}(p_i\leq t)\leq t$, $\forall\: t\in[0,1]$ and $i\in\mathcal{H}_0$, where $\mathcal{H}_0$ denotes the true null hypotheses, see for example, \cite{li2017accumulation, ramdas2019unified}.  Adaptive FDR control procedure tends to require stronger assumptions. \cite{tian2019addis} assumes that all the null $p$-values are {\it uniformly conservative}, i.e.,  
	$\mathbb{P}(p_i/\tau\leq t \mid p_i\leq\tau)\leq t$, $\forall\: t,\tau\in(0, 1)$. The AdaPT procedure proposed by \cite{lei2018adapt} assumes that the null $p$-values are {\it mirror conservative}:
\begin{equation}
   \label{eq: mirror-con}
	\mathbb{P}\left(p_i\in[a_1,a_2] \right)\leq \mathbb{P}\left(p_i\in[1-a_2, 1-a_1] \right), \quad \forall\: 0\leq a_1\leq a_2\leq 0.5. 
\end{equation}	
	Those assumptions on null $p$-values all cover the uniform distribution as a special case and hold under various scenarios, as discussed in the literature. Intuitively, mirror-conservatism allows us to control the quantity of small null $p$-values by referencing the number of large null $p$-values, thereby providing a way to control the FDR. It is important to note that mirror-conservatism doesn't automatically result in super-uniformity, and likewise, super-uniformity doesn't guarantee mirror-conservatism. Null $p$-values with a convex CDF or a monotonically increasing density are uniformly conservative, and such conservatism implies both super-uniformity and mirror-conservatism. This paper will build on the mirror conservative assumption to develop an adaptive differentially private FDR control procedure.

	%A novel procedure based on the optional-stopping criterion is recently proposed to control FDR, see \cite{storey2004strong}. Many recent works on FDR control extend the optional-stopping approach. For example, \cite{barber2015controlling} and \cite{barber2019knockoff} extend the sequential procedure to selective inference in regressions. Our work is mainly based on the \textit{Adaptive $p$-value Thresholding} (AdaPT) procedure proposed in \cite{lei2018AdaPT}, which is an iterative and interactive optional-stopping method. 
	
	%Our work modifies the AdaPT procedure with a differential privacy guarantee. Compared to the private BH procedure in \cite{dwork2021differentially}, our proposed private AdaPT can exploit auxiliary information to improve the power, and our proposed procedure has valid FDR control.

	\subsection{Related Work and Contributions}
	
	The most related work to our paper is the differentially private BH procedure proposed in \cite{dwork2021differentially}. \cite{dwork2021differentially} provides conservative bounds for $\mbox{FDR}_k:=\mathbb{E}[V/R\mid V\geq k]$, $k\geq 2$, and $\mbox{FDR}^k:=\mathbb{E}[V/R\mid R\geq k]$, $k\geq 1$. However, conservatism is unavoidable in the approach proposed in \cite{dwork2021differentially} due to the additional noise required for privacy. 
 
 %Moreover, the DP-BH procedure can control $\mbox{FDR}_k\leq C_k \alpha$ and $\mbox{FDR}^k\leq 1.1\alpha+2\sqrt{k\alpha}$, where $\alpha$ is the user-specified FDR level and $C_k$ is a constant larger than 1. The additional constant $C_k$ and $2\sqrt{k\alpha}$ are less satisfying because it inflates the desired FDR level. However, $C_k$  and $2\sqrt{k\alpha}$ are unavoidable in the approach proposed in \cite{dwork2021differentially} due to the additional noise required for privacy.
	
	In this paper, we propose an adaptive differentially private FDR control method that is able to control the FDR in the classical sense: $\mbox{FDR}\leq\alpha$, without any conditional component for the false discovery proportions or the constant term that inflates $\alpha$. Our work is based on a novel private $p$-value transformation mechanism that can protect the privacy of individual $p$-values while maintaining the mirror conservative assumption on the null $p$-values. By further developing a mirror peeling algorithm, we can define a filtration and apply the optimal stopping technique to prove that the proposed DP-AdaPT method controls FDR at any user-specified level $\alpha$ with finite samples. Theoretically, the proposed method provides a stronger guarantee on false discovery rate control compared to the differentially private BH method. Numerically, the proposed method works as well as the differentially private BH method when only the $p$-values are available for each test, and performs better when side information is available. The proposed method is also model-free when incorporating the side information for each hypothesis test. Lastly, the method is shown to only incur a small accuracy loss compared to the non-private AdaPT \citep{lei2018adapt} but at the same time reduces huge computation costs.
	
	This paper is organized as follows. Section 2 defines the basic concepts of differential privacy and briefly introduces the AdaPT procedure of FDR control. Section 3 provides the private $p$-value transformation mechanism, the definition of sensitivity for $p$-values, the DP-AdaPT algorithm, and the guaranteed FDR control. We demonstrate the numerical advantage of AdaPT through extensive simulations in Section 4 and conclude the paper with some discussion on future work in Section 5.

 \section{Preliminaries}
	
	\subsection{Differential Privacy}
	
	We first introduce the background for differential privacy. A dataset $\mathcal{S}=\{\boldsymbol{x}_1,\dots,\boldsymbol{x}_n\}$ is a collection of $n$ records, where $\boldsymbol{x}_i\in\mathcal{X}$ for $i=1,\dots,n$ and $\mathcal{X}$ is the domain of $\boldsymbol{x}$. The random variable $\{\boldsymbol{x}_i\}_{i=1}^{n}$ does not have to be independent. Researchers are usually concerned with certain statistics or summary information based on the dataset, denoted as $\mathcal{T}(\mathcal{S})$. For example, one might be interested in the sample mean, the regression coefficients,  or specific test statistics. When the dataset $\mathcal{S}$ is confidential and contains private individual information, researchers prefer to release a randomized version of $\mathcal{T}(\mathcal{S})$, which we denote as $\mathcal{M}\left(\mathcal{S}\right)$. A \textit{neighboring dataset} to $\mathcal{S}$ is denoted by $\mathcal{S}^{\prime}=\{\boldsymbol{x}_1^{\prime},\dots,\boldsymbol{x}_n^{\prime}\}$, with the requirement that only one index $j\in\{1,\dots,n\}$ satisfies that $\boldsymbol{x}_j\neq\boldsymbol{x}_j^{\prime}$. The classic $\left(\epsilon,\delta\right)$-Differential Privacy \citep{dwork2006calibrating} is defined as follows.
	
	\begin{defi}
		\label{def-ed-dp}
		A randomized mechanism $\mathcal{M}\left(\cdot\right)$ is $\left(\epsilon,\delta\right)$-differentially private for $\epsilon>0$ and $\delta>0$, if for all neighboring datasets $\mathcal{S}$ and $\mathcal{S}^{\prime}$, and any measurable set $E$,
		\begin{equation}
			\label{ed-dp}
			\mathbb{P}\left(\mathcal{M}\left(\mathcal{S}\right)\in E\right)\leq e^{\epsilon}\mathbb{P}\left(\mathcal{M}\left(\mathcal{S}^{\prime}\right)\in E\right)+\delta.
		\end{equation}
	\end{defi}
	When $\delta=0$, Definition \ref{def-ed-dp} is the {\it  pure differential privacy} and denoted by $\epsilon$-DP. When $\delta>0$, it is called the {\it approximate differential privacy}. The two neighboring datasets are treated as fixed, and the mechanism $\mathcal{M}\left(\cdot\right)$ contains randomness that is independent of the dataset and protects privacy. The set $E$ is measurable with respect to the random variable $\mathcal{M}\left(\cdot\right)$. In the definition, the two parameters $\epsilon$ and $\delta$ control the difference between the likelihood of $\mathcal{M}\left(\mathcal{S}\right)$ and $\mathcal{M}\left(\mathcal{S}^{\prime}\right)$. A small value of $\epsilon$ and $\delta$ indicates that the difference between the distribution of $\mathcal{M}\left(\mathcal{S}\right)$ and $\mathcal{M}\left(\mathcal{S}^{\prime}\right)$ is small and, as a result, using the outcome from the mechanism $\mathcal{M}$, one can hardly tell whether a single individual is included in the dataset $\mathcal{S}$. Thus, privacy is guaranteed with high probability for each individual in the dataset $\mathcal{S}$.
	
	%To provide privacy guarantee in $\mathcal{M}\left(\cdot\right)$, a natural solution is to perturb the original statistic $\mathcal{T}\left(\mathcal{S}\right)$ by some random noise. Those noises are usually independent random variables drawn from Laplace distribution or Gaussian distribution, with their noise calibrated to the \textit{sensitivity} of $\mathcal{T}\left(\mathcal{S}\right)$.% \caicomment{Define sensitivity}
	
	\cite{dong2021gaussian} proposed to formulate privacy protection as a hypothesis-testing problem for two neighboring datasets $\mathcal{S}$ and $\mathcal{S}^{\prime}$:
     \begin{equation}
      \label{test}
	H_0:\text{ the underlying dataset is }\mathcal{S},\quad\text{versus}\quad H_1:\text{ the underlying dataset is }\mathcal{S}^{\prime}.   
     \end{equation}
	Let $\boldsymbol{x}_j$ denote the only individual in $\mathcal{S}$, but not in $\mathcal{S}^{\prime}$. Accepting the null hypothesis implies identifying the presence of $\boldsymbol{x}_j$ in the dataset $\mathcal{S}$, and rejecting the null hypothesis implies identifying the absence $\boldsymbol{x}_j$. Thus privacy can be interpreted by the power function of testing (\ref{test}). Specifically, the $\mu$-Gaussian differential privacy is defined as a test that is at least as hard as distinguishing between two normal distributions $N(0,1)$ and $N(\mu, 1)$ based on one random draw of the data. For the readers' convenience, we rephrase the formal definition from \cite{dong2021gaussian}.
	
	%By theorem 2.1 in \cite{kairouz2015composition}, any test method at significance level $\alpha\in\left(0,1\right)$, based on the output of a $\left(\eps,\delta\right)$-DP mechanism, has power non-greater than $e^{\eps}\alpha+\delta$. The composition of an $\left(\eps_1,\delta_1\right)$-DP mechanism and an $\left(\eps_2,\delta_2\right)$-DP mechanism is $\left(\eps_1+\eps_2,\delta_1+\delta_2\right)$-DP, and leads to an upper bound $e^{\eps_1+\eps_2}\alpha+\delta_1+\delta_2$ of power, which is not tight \citep{dong2021gaussian}. To handle the tradeoff between size and power, a recent version of differential privacy is proposed by \cite{dong2021gaussian}.
	
	%Of particular interest to this paper is the recent work on Gaussian differential privacy proposed in \cite{dong2021gaussian}. By taking a hypothesis testing perspective, \cite{dong2021gaussian} proposed a new definition of privacy that can only rely on one parameter to quantify privacy and is shown to maintain a tight privacy guarantee under multiple compositions of private mechanisms. Thus it is particularly useful for statistical methods that require multiple or iterative operations of the data. 
	
	\begin{defi}[Gaussian Differential Privacy]
		\begin{enumerate}
			\item 	A mechanism $\mathcal{M}$ is $f$-\textit{differential private} ($f$-DP) if any $\alpha$-level test of (\ref{test}) has power function $\beta(\alpha)\leq 1-f(\alpha)$, where $f(\cdot)$ is a convex, continuous, non-increasing function satisfying $f(\alpha)\leq 1-\alpha$ for all $\alpha\in[0,1]$.
			\item  	A mechanism $\mathcal{M}$ is $\mu$-\textit{Gaussian Differential Privacy} ($\mu$-GDP) if $\mathcal{M}$ is $G_\mu$-DP, where $G_{\mu}\left(\alpha\right)=\Phi\left(\Phi^{-1}\left(1-\alpha\right)-\mu\right)$ and $\Phi\left(\cdot\right)$ is the cumulative distribution function of $N(0,1)$.
		\end{enumerate}
	\end{defi}
	
	The new definition has several advantages. For example, privacy can be fully described by a single mean parameter of a unit-variance Gaussian distribution, and this makes it easy to describe and interpret the privacy guarantees. The privacy definition is shown to maintain a tight privacy guarantee under multiple compositions of private mechanisms. Thus, it is particularly useful for statistical methods that require multiple or iterative operations of the data. We will use the definition of $\mu$-GDP throughout the rest of this paper. The proposed method can be easily extended to the classic $\left(\epsilon,\delta\right)$-DP by Corollary 1 in \cite{dong2021gaussian}.
	
	\subsection{Adaptive False Discovery Rate Control}
	
	In this subsection, we introduce the AdaPT procedure proposed by \cite{lei2018adapt}, which is described in Algorithm 1 for completeness. Assume we have $p$-values $p_i$ and  side information $\boldsymbol{x}_i$ for each hypothesis $H_i$, $i=1,\dots,n$. The procedure contains an iterative update of covariate-specific thresholds. At each step $t=0,1,\dots$, a rejection threshold $s_t\left(\boldsymbol{x}\right)$ is decided based on the covariate $\boldsymbol{x}$. Let $R_t=\left|\{i:p_i\leq s_t\left(\boldsymbol{x}_i\right)\}\right|$, $A_t=\left|\{i:p_i\geq 1-s_t\left(\boldsymbol{x}_i\right)\}\right|$ and the estimated false discovery rate $\widehat{\text{FDR}}=\left(1+A_t\right)/\left(R_t\vee 1\right)$. If $\widehat{\text{FDR}}\leq\alpha$, then we stop and reject all the $H_i$ with $p_i\leq s_t\left(\boldsymbol{x}_i\right)$. Otherwise, we update the thresholds $s_{t+1}\preceq s_t$, where $s_{t+1}\preceq s_t$ denotes $s_{t+1}\left(\boldsymbol{x}\right)\leq s_t\left(\boldsymbol{x}\right)$ for every $\boldsymbol{x}$ in the domain of $s\left(\cdot\right)$. The information that are used to update $s_{t+1}$ contains $A_t$, $R_t$ and $\left(\boldsymbol{x}_i,p_{pm,i}\right)_{i=1}^n$, where
	$$p_{pm,i}=
	\begin{cases}
		p_i& s_t\left(\boldsymbol{x}_i\right)<p_i<1-s_t\left(\boldsymbol{x}_i\right)\\
		\{p_i,1-p_i\} &\text{otherwise,} \\ 
	\end{cases}$$
	is partially masked $p$-values and the subscript $pm$ denotes ``partially masked". The partially censored $p$-values restrict the analyst's knowledge and enables the application of the optional stopping technique widely used in the FDR literature \citep{storey2004strong, barber2015controlling, li2017accumulation}. In this paper, we will develop a mirror peeling algorithm that builds on this novel technique and prove the desired FDR guarantee with differential privacy.
	
	%In the remaining paper, we develop an adaptive FDR control procedure with privacy guarantee.
	
	\begin{algorithm}
		\caption{AdaPT \citep{lei2018adapt}}
		\begin{algorithmic}
			\Require $\{\boldsymbol{x}_i,p_i\}_{i=1}^n$, initialization $s_0$, target FDR level $\alpha$.
			\ForAll  {$t=0$ to $\dots$}
			\State $\widehat{\text{FDR}}\leftarrow\frac{1+A_t}{R_t\vee 1}$
			\If {$\widehat{\text{FDR}}\leq\alpha$}
			\State Reject $\{H_i:p_i\leq s_t\left(\boldsymbol{x}_i\right)\}$
			\State Return $s_t$
			\EndIf
			\State $s_{t+1}\leftarrow\text{Update}\left[\left(\boldsymbol{x}_i,p_{pm,i}\right)_{i=1}^{n},A_t,R_t,s_t\right]$
			\EndFor 
		\end{algorithmic}
	\end{algorithm}

 \section{Methodology}
	
	Consider $n$ hypotheses $H_1,H_2,\dots, H_n$, and researchers can observe side information $\boldsymbol{x}_i$ and estimate a $p$-value for each hypothesis $H_i$. In this section, we aim to develop a differentially private algorithm that protects the privacy of individual $p$-values and controls the FDR at the same time. The analysis does not rely on the threshold model $s\left(\boldsymbol{x}\right)$, and is model-free. We assume that auxiliary information $\boldsymbol{x}$ is public and is not subject to privacy concerns. This assumption is reasonable because the auxiliary information is usually from scientific knowledge or previous experiments. With a specific model for the threshold function $s\left(\boldsymbol{x}\right)$, the proposed method can also be easily extended to further protect the privacy of $\boldsymbol{x}$. 
	
	\subsection{Private p-value}
	
	Following \cite{lei2018adapt}, we assume that all the null $p$-values satisfy the mirror-conservative property as defined in (\ref{eq: mirror-con}). We first propose a novel differentially private mechanism on the individual $p$-values that protects privacy while still satisfying the mirror-conservative property. This is a crucial property because it helps us avoid the traditional technique in the differential privacy literature (e.g., \cite{dwork2021differentially}) that derives conservative error bounds on the noise added for privacy.
	
	The proposed mechanism is based on the quantile function and cumulative distribution function of some symmetric distributions. Specifically, let $U\in\mathbb{R}^+$ be the boundary and can possibly take the value of $\infty$. Let $g(\cdot):\left(-U,U\right)\to\mathbb{R}^+$ be an integrable function satisfying the following conditions:
	\begin{enumerate}
		\item Non-negative: $g\left(x\right)\geq 0$ for $x\in\left(-U,U\right)$, $g\left(x\right)= 0$ for $x\notin\left(-U,U\right)$ and the measure of the set $\{x\in\left(-U,U\right):g\left(x\right)=0\}$ is zero with respect to the measure $\mu$ on $\mathbb{R}$;
		\item Symmetric: $g\left(x\right)=g\left(-x\right)$ for $x\in\left(-\infty,\infty\right)$;
		\item Unity: $\int_{-\infty}^{\infty}g\left(x\right)\mu \left(dx\right)=1$.
	\end{enumerate}
	The primitive function of $g\left(\cdot\right)$ is denoted by $G\left(x\right)=\int_{-\infty}^{x}g\left(x\right)\mu \left(dx\right)$. The function $g\left(\cdot\right)$ can be viewed as a symmetric probability density function, and the function $G\left(\cdot\right)$ can be viewed as a strictly increasing cumulative distribution function. The function $G\left(\cdot\right)$ is a one-to-one mapping from $\left(-U,U\right)$ to $\left[0,1\right]$, which guarantees the existence of a quantile function $G^{-1}\left(\cdot\right)$.  We will use $G^{-1}(\cdot)$ and $G(\cdot)$ to transform the $p$-values. When the distribution of $p$-value is continuous, the measure $\mu$ can be chosen as the Lebesgue measure.
	
	\begin{theorem}\label{lem: noisy_p_value_mc}
		Let the $p$-value $p$ be mirror-conservative, and $Z$ be an independent Gaussian random variable with mean zero and positive variance. Then the noisy $p$-value
  \begin{equation}
  \label{eq:noisy_p_value}
     \tilde{p}:= G\big\{G^{-1}\left(p\right)+Z\big\}
  \end{equation} is also mirror-conservative.
	\end{theorem}

	\begin{figure}[ht]
		\centering
		\includegraphics[width=0.65\textwidth]{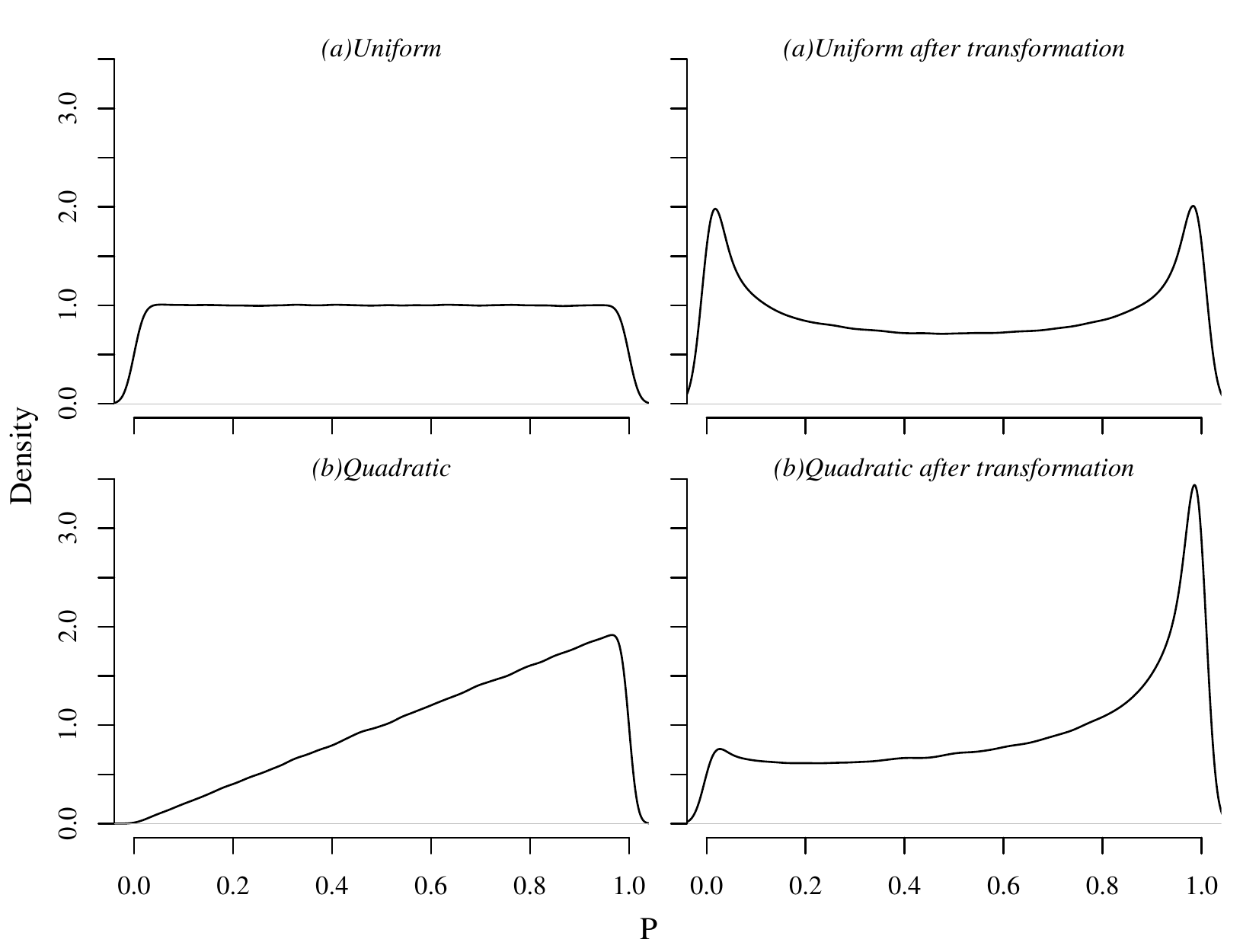}
		\caption{Empirical density estimate of both the original and the transformed $p$-values. Case a: the null $p$-values follow the uniform distribution; case b: the null $p$-values have density $f\left(p\right)=2p$ for $p\in\left[0,1\right]$. In both cases, $G(\cdot)=\Phi(\cdot)$ and $Z$ follows the standard normal distribution. The density curves are estimated by $\numprint{100000}$ samples.}
		\label{fig:noisy_mirror}
	\end{figure}
	
	The proof of Theorem \ref{lem: noisy_p_value_mc} is provided in the appendix. 
	Although Theorem \ref{lem: noisy_p_value_mc} is based on the Gaussian noise, one can easily extend the theory to the case where $Z$ follows Laplace distribution which is frequently used in the classic $\left(\epsilon,\delta\right)$-DP setting, see the discussions in Theorem \ref{thm: dpadapt_ed}. We provide two illustrative examples in  Figure \ref{fig:noisy_mirror}, where the empirical density estimates of both the original and the transformed $p$-values are plotted. On the top row, we show that when the original $p$-values follow the uniform distribution, the transformed noisy $p$-values are symmetric around $0.5$. On the bottom row, we show that when the original $p$-values are strictly stochastically larger than the uniform distribution, the transformed noisy $p$-values also tend to concentrate on the right side of the curve. Figure \ref{fig:noisy_mirror} visually demonstrates that the two most commonly encountered $p$-values are able to preserve the mirror conservative property.
	Note that in Figure \ref{fig:noisy_mirror}, we implemented the standard normal distribution to transform the $p$-values:  $G(\cdot)=\Phi(\cdot)$ and $g(\cdot) = \phi(\cdot)$.  The normal density $ \phi(\cdot)$ is monotone on either the positive or negative part of the horizontal axis. As a result, the transformed noisy $p$-values will concentrate on the two endpoints $0$ and $1$. 
	
	It is also interesting to note that Theorem \ref{lem: noisy_p_value_mc} does not necessarily hold for the other conservative assumptions on $p$-values, such as the super uniform assumption or the uniformly conservative assumption as discussed in the introduction. One can easily construct a counterexample that violates the requirements. The mirror conservative condition is the most appropriate in the sense of preserving differential privacy. Throughout the rest of the paper, we will use $\tilde{p}$ to denote the noisy $p$-values defined in (\ref{eq:noisy_p_value}). 
	
	\subsection{Sensitivity of p-values}
	
	The transformation in Theorem \ref{lem: noisy_p_value_mc} is useful in preserving the mirror-conservative property, but we need to calibrate the variance of noise $Z$ in order to provide privacy guarantee with minimal losses on accuracy. In this section, we provide a definition for the \textit{sensitivity} of $p$-values that directly fits into the framework of the transformation in Theorem \ref{lem: noisy_p_value_mc}. We begin with defining the \textit{sensitivity} for any deterministic real-valued functions. The definition provides an upper bound of the difference in the outcome due to the change of one item in the dataset. 
	
	\begin{defi}
		Let $g:\mathcal{X}^n\to\mathbb{R}$ be a deterministic function from the data set $\mathcal{S}$ to $\mathbb{R}$. The \textit{sensitivity} of $g(\cdot)$ is defined by
		$$\Delta\left(g\right):=\sup_{\text{$\forall$ }\mathcal{S},\mathcal{S}^{\prime}}\left\|g\left(\mathcal{S}\right)-g\left(\mathcal{S}^{\prime}\right)\right\|,$$
		where $\mathcal{S}^{\prime}$ is a neighboring dataset of $\mathcal{S}$, and $\|\cdot\|$ is the Euclidean norm.
	\end{defi}
	
	In this paper, we consider the case where the $p$-value is estimated from a non-randomized decision rule, where the $p$-value is a deterministic real-valued function of the data. However, due to its nature, the relative change of a $p$-value on two neighboring datasets is usually very small. Thus directly adding noise to $p$-values may easily overwhelm the signals and lead to unnecessary power losses. For example, \cite{dwork2021differentially} controls the sensitivity of $p$-value based on a truncated log transformation, and the truncation parameter has to be carefully tuned to make a tradeoff between privacy and accuracy. 
	
	In this paper, we define the sensitivity by considering a transformed $p$-value based on the function $G^{-1}(\cdot)$. The transformation is motivated by the fact that the $p$-values are usually obtained based on the limiting null distribution of the test statistics, which are, in most cases, asymptotically normal. For example, the $p$-value of a one-sided mean test is the quantile function of a normal distribution evaluated at the sample mean. We provide the formal definition as follows.
	
	\begin{defi}[Sensitivity]
		\label{def_sensi}
		The sensitive of $p$-value function $p$ is $\Delta_G$ if for all neighboring dataset $\mathcal{S}$ and $\mathcal{S}^{\prime}$,
		$$\sup_{\text{$\forall$ }\mathcal{S},\mathcal{S}^{\prime}}\left\|G^{-1}\left\{ p\left(\mathcal{S}\right)\right\}-G^{-1}\left\{ p\left(\mathcal{S}^{\prime}\right)\right\}\right\|\leq\Delta_G.$$
	\end{defi}
	
	The choice of $G\left(\cdot\right)$ function in the Definition \ref{def_sensi} is flexible. For example, if the density of the test statistics under the null hypothesis is symmetric, then one can choose $G\left(\cdot\right)$ as the CDF of the test statistic. We provide the following examples where the desired $\Delta_G$ is calculated under mild conditions.
	
	\begin{example}
		\label{exam:one-side}
		Assume $X_1, X_2,\dots, X_n$ are i.i.d. random variables with mean $\beta$, variance $1$, and are uniformly bounded by $M$. To test null $H_0:\beta\geq 0$ against alternative $H_1:\beta<0$, we use the statistics $T=\sum_{i=1}^{n}X_i/\sqrt{n}$. With a large sample size,  the $p$-value is estimated as $\widehat{p}=\Phi\left(T\right)$. The $G(\cdot)$ function in Definition \ref{def_sensi} can be chosen as $\Phi(\cdot)$, and $\Delta_G=2M/\sqrt{n}$.
	\end{example}
	
	\begin{example}
		\label{exam:two-side}
		Assume $X_1, X_2,\dots, X_n$ are i.i.d. random variables with mean $\beta$, variance $1$, and are uniformly bounded by $M$. To test null $H_0:\beta= 0$ against alternative $H_1:\beta\neq 0$, we also use the statistics $T=\sum_{i=1}^{n}X_i/\sqrt{n}$. With a large sample size,  the $p$-value is estimated as $\widehat{p}=2\Phi\left(-\left|T\right|\right)$.  Let $\text{supp}\{g\left(x\right)\}\subset \left[-M,M\right]$ with bounded density and $g\left(M\right)=g\left(-M\right)>0$, $G\left(x\right)=\int_{-\infty}^{x}g\left(x\right)dx$. For example, $g(\cdot)$ can be the density of a truncated normal distribution supported on $[-M, M]$. Then $\Delta_G=2MC/\sqrt{n}$, where $C$ is a constant. Detailed proofs are provided in the appendix. 
	\end{example}

	\begin{example}
		\label{exam:chi-sq}
		Assume $X_1, X_2,\dots, X_n$ are i.i.d. random variables with mean $\beta $, variance $\sigma^2$, and are uniformly bounded by $M$. Let $h\left(X_1,X_2\right)=X_1X_2$ and
		$$U_n ={\binom{n}{2}}^{-1}\sum_{i,j} h\left(X_i, X_j \right),$$
		Then we have
		$$\frac{nU_n}{\sigma^2}\stackrel{d}{\to}\chi^2_2-1,$$
		as $n\to\infty$. In this case, the $p$-value for testing the null $H_0:\beta= 0$ against the alternative $H_1:\beta\neq0$ is based on the statistics $T=nU_n/\sigma^2+1$. The $G(\cdot)$ function in Definition \ref{def_sensi} can be chosen as $\Phi(\cdot)$ and the sensitivity is bounded by $$\Delta_G=M^2/n C_1+\frac{C_2}{1/2-\delta}\left(M^2/n\right)^{1/2-\delta},$$ where $C_1$ and $C_2$ are constants, and $0<\delta<1/2$. Detailed proofs are provided in the appendix.
	\end{example}
	
	The sensitivity in Example \ref{exam:one-side} is tight, due to the nature of the one-sided test and normal transformation of the $p$-value. In Example \ref{exam:two-side}, we used the truncated normal distribution to perform the transformation to simplify the technical calculation. In Example \ref{exam:chi-sq}, the $\Delta_G$ is approximately the square root of the original sensitivity $\Delta$ because of the imperfect match between the tail of the transformation function (normal distribution) and the tail of the $\chi^2$ distribution. In fact, the normal transformation, i.e., $G(\cdot) = \Phi(\cdot)$ works well for most cases, as we will show in the numerical studies.

	\subsection{The Differentially Private AdaPT algorithm}
	In this section, we propose the DP-AdaPT algorithm. % In the remaining paper, we assume the sensitivity of all $G^{-1}\left(p\right)$ is bounded by $\Delta$. 
	We begin the discussion by introducing the Gaussian mechanism. To report a statistic $\mathcal{T}\left(\mathcal{S}\right)$ with the privacy guarantee, \textit{Gaussian mechanism} adds noise to the target statistics $\mathcal{T}\left(\mathcal{S}\right)$, with the scale of noise calibrated according to the sensitivity of $\mathcal{T}\left(\mathcal{S}\right)$. We summarize some appealing properties of the Gaussian mechanism in Lemma \ref{lem: gaussian mecha}.

	\begin{lemma}\label{lem: gaussian mecha}
		The Gaussian Mechanism has the following properties \citep{dong2021gaussian}:
		\begin{enumerate}
			\item GDP guarantee. The Gaussian mechanism that outputs
			$\mathcal{M}\left(\mathcal{T}\right)=\mathcal{T}\left(\mathcal{S}\right)+Z$
			preserves $\mu$-GDP, where $Z$ is drawn independently from $\mathcal{N}\left(0,\text{sene}\left(\mathcal{T}\right)^2/\mu^2\right)$ and $\text{sene}\left(\mathcal{T}\right)$ is the sensitivity of $\mathcal{T}$ defined in the Definition 3.
			\item Composition. Let $\mathcal{M}_1$ and $\mathcal{M}_1$ be two algorithms that are $\mu_1$-GDP and $\mu_2$-GDP, respectively. The composition algorithm $\mathcal{M}_1\circ\mathcal{M}_2$ is $\sqrt{\mu_1^2+\mu_2^2}$-GDP.
			\item Post-processing. Let $f\left(\cdot\right)$ be a deterministic function and $\mathcal{M}$ be a $\mu$-GDP algorithm. Then the post-processed algorithm $f\circ\mathcal{M}$ is $\mu$-GDP.
		\end{enumerate}
	\end{lemma}
	
	%The type \Rmnum{1} error $\alpha$ and power trade-off of the composition algorithm $\mathcal{M}_1\circ\mathcal{M}_2$ is defined by the function $\Phi\left(\Phi^{-1}\left(1-\alpha\right)-\sqrt{\mu_1^2+\mu_2^2}\right)$. The upper bound of power is tighter than the composition of $\left(\eps,\delta\right)$-DP algorithm. 
	
	In multiple testing, a common scenario is that the number of hypotheses is very large. If we report all the $p$-values under the private parameter $\mu$, the standard deviation of the noise is proportional to the square root of the number of hypotheses by the composition lemma. Thus, in large-scale hypothesis testing, reporting all $p$-values adds very large noise to the signal and weakens the power of tests. To overcome the difficulty, the first step of our algorithm is to select a subset of $p$-values with more potential to be rejected, and the second step is to report the subset of $p$-values with a privacy guarantee. It is also common in real practice that the true signals are only a small subset of the total hypotheses. For example, only a few genes are truly related to the phenotype of interest. 
	The following report noisy min algorithm \citep{dwork2021differentially} builds the foundation of the selection algorithm.
	
	\begin{algorithm}
		\caption{The Report Noisy Min Algorithm}
		\label{algorithm-report-min}
		\begin{algorithmic}
			\Require $p$-values $p_1,\dots,p_n$ each with sensitivity at most $\Delta$, privacy parameter $\mu$.
			\ForAll  {$j=1$ to $n$}
			\State set $\tilde{f}_j=G[G^{-1}(p_j\left(\mathcal{S}\right))+Z_j]$, where $Z_j$ is an independent sample from normal distribution with mean $0$ and variance $8\Delta^2/\mu^2$;
			\EndFor 
			\State
			return $j^*=\arg\min_j\tilde{f}_j$ and $\tilde{p}_{j^{*}}:=G[G^{-1}(p_{j^*}(\mathcal{S}))+Z]$, where $Z$ is an independent, afresh drawn sample from normal distribution with mean $0$ and variance $8\Delta^2/\mu^2$.
		\end{algorithmic}
	\end{algorithm}
	
	\begin{lemma}
		\label{report_nosiy_min_al_lemma}
		Report noisy min algorithm, as detailed in Algorithm \ref{algorithm-report-min}, is $\mu$-GDP.
	\end{lemma}
	
	A traditional way to select the most important signals is the peeling algorithm \citep{cai2021cost, dwork2021differentially}, which repeats the report noisy min algorithm for a fixed number of times. However, the peeling algorithm creates complex dependent structures among the selected $p$-values, and further complicates the analysis of FDR control. In fact, we believe it is the main issue in \cite{dwork2021differentially} that prevented the authors from bounding the classic FDR criterion instead worked on the conditional quantity $\mbox{FDR}_k$, with $k\geq 2$.
	
	In this paper, we propose a novel mirror peeling algorithm that perfectly suits the situation of adaptive FDR control. The selection procedure is based on the partially masked $p$-values and simultaneously selects both the largest and the smallest $p$-values. The largest $p$-values will be used to estimate the false discovery proportion as the control, which is a widely used technique in the multiple testing literature.

	\begin{algorithm}
		\caption{The Mirror Peeling Algorithm}
		\label{al:mirror_peeling_lemma}
		\begin{algorithmic}
			\Require $p$-values $p_1\dots,p_n$ each with sensitivity at most $\Delta$, privacy parameter $\mu$, size $m$.
                \State let $\mathcal{S}=\{1,\dots,n\}$ be the index set of $p$-values;
			\ForAll  {$j=1$ to $m$}
			\State 
                let $i_j$ be the returned index of the report noisy min algorithm applied to partially masked $p$-values $$\{\min\left(p_i,1-p_i\right)\}_{i\in\mathcal{S}},$$
		   with $\mu=\mu/\sqrt{m}$; 
   \State let $\tilde{p}_{i_j}$ be the noisy $p_{i_j}$ defined in (\ref{eq:noisy_p_value}) with $Z$ following independent normal noise with mean $0$ and variance $8m\Delta^2/\mu^2$;
			\State
			Update $\mathcal{S} = \mathcal{S}\backslash\{i_j\}$;
			\EndFor 
			\State
			return $\bigg\{\left(i_1,\tilde{p}_{i_1}\right),\dots,\left(i_{m},\tilde{p}_{i_{m}}\right)\bigg\}$.
		\end{algorithmic}
	\end{algorithm}
	
	\begin{lemma}
		\label{mirror_peeling_lemma}
		The mirror peeling algorithm, as presented in Algorithm \ref{al:mirror_peeling_lemma}, is $\mu$-GDP.
	\end{lemma}
	
	The size $m$ in the mirror peeling algorithm denotes the number of selected $p$-values. In practice, we suggest choosing a slightly large $m$ to prevent potential power loss. Now we are ready to state the DP-AdaPT procedure in Algorithm \ref{al:DP-AdaPT}, which controls the FDR at a user-specified level $\alpha$ with guaranteed privacy. Theorem \ref{thm: dpadapt} follows directly by Lemma \ref{report_nosiy_min_al_lemma} and \ref{mirror_peeling_lemma} and the post-processing property of GDP algorithms.
	
	\begin{algorithm}
		\caption{The DP-AdaPT Algorithm}
		\label{al:DP-AdaPT}
		\begin{algorithmic}
			\Require target FDR level $\alpha$, $\{\boldsymbol{x}_i,p_i\}_{i=1}^n$, $p$-values with sensitivity at most $\Delta$, privacy parameter $\mu$ and size $m$.
			\State Apply the mirror peeling algorithm and obtain $\bigg\{\left(i_1,\tilde{p}_{i_1}\right),\dots,\left(i_{m},\tilde{p}_{i_{m}}\right)\bigg\}$.
			\ForAll  {$t=0,1, \dots$}
			\State  Let $R_t=\left|\{i_j,j=1,\dots,m:\tilde{p}_{i_j}\leq s_t\left(\boldsymbol{x}_{i_j}\right)\}\right|$,\\ $A_t=\left|\{i_j,j=1,\dots,m:\tilde{p}_{i_j}\geq 1-s_t\left(\boldsymbol{x}_{i_j}\right)\}\right|$
			\State $\widehat{\text{FDR}}\leftarrow\frac{1+A_t}{R_t\vee 1}$
			\If {$\widehat{\text{FDR}}\leq\alpha$}
			\State Reject $\{H_{i_j}:\tilde{p}_{i_j}\leq s_t\left(\boldsymbol{x}_{i_j}\right)\}$;
			\State Return $s_t$;
			\EndIf
			\State $s_{t+1}\leftarrow\text{Update}\left[\left(\boldsymbol{x}_{i_j},\tilde{p}_{pm,i_j}\right)_{j=1}^{m},A_t,R_t,s_t\right]$, where 
			$$\tilde{p}_{pm,i_j}=
			\begin{cases}
				\tilde{p}_{i_j}& s_t\left(\boldsymbol{x}_i\right)<\tilde{p}_{i_j}<1-s_t\left(\boldsymbol{x}_i\right)\\
				\{\tilde{p}_{i_j},1-\tilde{p}_{i_j}\} &\text{otherwise.} \\ 
			\end{cases}$$
			\EndFor 
		\end{algorithmic}
	\end{algorithm}
	
	\begin{theorem}\label{thm: dpadapt}
		The DP-AdaPT algorithm described in Algorithm \ref{al:DP-AdaPT} is $\mu$-GDP.
	\end{theorem}

   For completeness of the discussion, we provide a classic $\left(\epsilon,\delta\right)$-private version of the proposed DP-FDR control algorithm. By the relation between $\mu$-GDP and $\left(\epsilon,\delta\right)$-DP, the Algorithm \ref{al:DP-AdaPT} is $\left(\epsilon,\delta\right)$-DP for $\forall\:\epsilon>0$ and $\delta=\Phi\left(-\epsilon/\mu+\mu/2\right)-e^{\epsilon}\Phi\left(-\epsilon/\mu-\mu/2\right)$. With pre-specified private parameters $\epsilon$ and $\delta$, we proposed a modification of Algorithm \ref{al:DP-AdaPT} which uses the Laplace mechanism in The Report Noisy Min Algorithm. Theorem \ref{thm: dpadapt_ed} shows the proposed modified algorithm is $\left(\epsilon,\delta\right)$-DP.

 \begin{theorem}\label{thm: dpadapt_ed}
	Given private parameters $\epsilon\leq 0.5$, $\delta\leq 0.1$, the peeling size $m\geq 10$ and sensitivity at most $\Delta$, the DP-AdaPT algorithm described in Algorithm \ref{al:DP-AdaPT} with $Z_j$ and $Z$ in Algorithm \ref{algorithm-report-min} following Laplace noise of scale $\lambda=\Delta\sqrt{10m\log\left(1/\delta\right)}/\epsilon$ is $\left(\epsilon,\delta\right)$-differentially private.
	\end{theorem}

%We remark that the Theorem \ref{thm: dpadapt_ed} is similar to Theorem 3 in \cite{dwork2021differentially}.

Although, the FDR control procedure of the proposed DP-AdaPT method is very different from the BH method used by \cite{dwork2021differentially}, the privacy is protected by a similar procedure-the peeling mechanism. With the same sensitivity parameter $\Delta$, the peeling size $m$, and privacy parameters $\left(\epsilon,\delta\right)$, our modified DP-AdaPT procedure uses the same level of noise as the DP-BH procedure proposed by \cite{dwork2021differentially}. With the same noise level, our proposed method is superior to the DP-BH method in terms of the exact valid FDR control and the higher power of detecting the true nulls.
 
	\subsection{FDR Control}
	
	There are a few challenges in deriving the FDR bound for differentially private algorithms. Firstly, the privacy-preserving procedure is required to be randomized with noise independent of the data. However, most classic FDR procedures implement fixed thresholds to decide the rejection regions and are unsuitable for noisy or permuted private $p$-values. Secondly, the mirror peeling algorithm creates complicated dependence structures among the selected $p$-values. Classic tools in the literature that are used for proving FDR control crucially rely on the independence assumption or the positive dependence assumptions on the $p$-values, thus become inapplicable for differential private algorithms. Thirdly, without the martingale technique \citep{storey2004strong}, it is in general difficult to derive finite sample results with differential privacy. In fact, it is still unclear how to obtain valid finite sample FDR control for the differential private BH procedure. 	The DP-BH algorithm proposed in \cite{dwork2021differentially} addressed the challenges by conservatively bounding the noise and derived the upper bound for an unusual conditional version of FDR, i.e., $\text{FDR}_k:=\mathbb{E}\left[V/R;V\geq k\right]$. 
	
	In this paper, we prove that the DP-AdaPT algorithm controls the FDR in finite samples. Our proof adopted the similar optional stopping argument in the multiple testing literature \citep{storey2004strong, barber2015controlling, li2017accumulation, lei2018adapt}. We show that the adaptive procedure and the mirror conservative assumption work perfectly with the additional noise required to protect privacy. By only using partial information in the mirror peeling algorithm, we can construct a filtration and apply the martingale technique. 
 
 We first introduce the notations. For each hypothesis $H_i$, we observe $p$-value $p_i$ and auxiliary information $\boldsymbol{x}_i$. Given a pre-specified sparsity level $m\leq n$, the DP-AdaPT algorithm first applies the mirror peeling algorithm, and we use $\mathcal{P}$ to denote the index returned by the mirror peeling algorithm.	Let $\mathcal{F}_t$ for $t=0,\dots,$ denote the filtration generated by all information available to the analyst at step $t$:
	$$\mathcal{F}_t=\sigma\left(\{\boldsymbol{x}_i,\tilde{p}_{pm,t,i}\}_{i\in\mathcal{P}},A_t,R_t\right)$$
	where 
	$$\tilde{p}_{pm,t,i}=
	\begin{cases}
		\tilde{p}_i& s_t\left(\boldsymbol{x}_i\right)<\tilde{p}_i<1-s_t\left(\boldsymbol{x}_i\right)\\
		\{\tilde{p}_i,1-\tilde{p}_i\} &\text{otherwise.} \\ 
	\end{cases}$$
	The initial $\sigma$ field is defined as $\mathcal{F}_{-1}=\sigma\left(\{\boldsymbol{x}_i,\{\tilde{p}_{i},1-\tilde{p}_{i}\}\}_{i\in\mathcal{P}}\right)$. The two updating thresholds principles: $s_{t+1}\preceq s_{t}$ and $s_{t+1}\in\mathcal{F}_{t}$, ensure that the $\{\mathcal{F}_t\}_{t=-1}^{\infty}$ is a filtration, i.e., $\mathcal{F}_{t}\subset\mathcal{F}_{t+1}$ for $t\geq -1$. 
	
	\begin{theorem}
		\label{thm:DP-AdaPT}
		Assume that all the null $p$-values are independent of each other and of all the non-null $p$-values, and the null $p$-values are mirror-conservative. The DP-AdaPT procedure controls the FDR at level $\alpha$.
	\end{theorem}
	
	Theorem \ref{thm:DP-AdaPT} has several important implications. First of all, it can control the FDR at a user-specified level $\alpha$ with differential privacy guarantee, while the existing DP-BH method fails. Secondly, due to the definition of filtration and the application of martingale tools, the DP-AdaPT can control the FDR for a finite number of tests. And lastly, when the side information is available to the hypothesis, the DP-AdaPT shares a similar property to the original AdaPT and can borrow the side information in a model-free sense to increase power. We demonstrate the numerical utilities in the next section.

\subsection{The Two-groups Working Model and Selection Procedure}

Our proposed DP-AdaPT procedure successfully controls FDR regardless of the strategy used in the threshold updating. In other words, it also enjoys the model-free property. But it is still important to provide a practical and powerful solution to update the threshold. \cite{lei2018adapt} proposed a two-group working model and use the local false discovery rate as the threshold. In this subsection, we illustrate a greedy procedure based on a two-group working model. Our procedure is similar to the method proposed by \cite{lei2018adapt} but has a simple illustration.

We begin with the working model specification. Assume that the distribution of hypothesis indicator $H_i$ given side information $\boldsymbol{x}_i$ follows Bernoulli distribution with probability $\pi\left(\boldsymbol{x}_i\right)$, $H_i\mid\boldsymbol{x}_i\sim\mbox{Bernoulli}\left(\pi\left(\boldsymbol{x}_i\right)\right)$, where $H_i=1$ if the $i$th hypothesis is true and $H_i=0$ otherwise. The distribution of observed $p$-value $p_i$ given $H_i$ and $\boldsymbol{x}_i$ satisfies:
\begin{equation*}
    p_i\mid H_i,\boldsymbol{x}_i\sim\begin{cases} f_0\left(p\mid\boldsymbol{x}_i\right) & \mbox{if } H_i=0 \\ f_1\left(p\mid\boldsymbol{x}_i\right) & \mbox{if } H_i=1 \end{cases}.
\end{equation*}
In addition, we assume the data $\{\left(\boldsymbol{x}_i,p_i,H_i\right)\}_{i=1}^{n}$ is mutually independent with $\{H_i\}_{i=1}^{n}$ missing or being unobserved. The $\pi\left(\cdot\right),f_0\left(\cdot\mid\boldsymbol{x}\right)$ and $f_1\left(\cdot\mid\boldsymbol{x}\right)$ are unknown functions and can be estimated by any user-specified methods. \cite{lei2018adapt} suggested using exponential families to model the $\pi\left(\cdot\right),f_0\left(\cdot\mid\boldsymbol{x}\right)$ and $f_1\left(\cdot\mid\boldsymbol{x}\right)$. As mentioned by \cite{lei2018adapt}, the model $\{\pi\left(\cdot\right),f_0\left(\cdot\mid\boldsymbol{x}\right),f_1\left(\cdot\mid\boldsymbol{x}\right)\}$ is not identifiable, and we use uniform distribution as the working model for the null hypothesis, $f_0\left(p\mid\boldsymbol{x}\right)=1$ for $p\in\left[0,1\right]$.

At the $t$-th iteration with available information $\mathcal{F}_t$, the first step is to fit the model using the data $\{\boldsymbol{x}_i,\tilde{p}_{pm,t,i}\}_{i\in\mathcal{P}}$. The complete log-likelihood at $t$-th iteration is
\begin{align}
l_t\left(\pi\left(\cdot\right),f_1\left(\cdot\right)\right)&=\sum_{i\in\mathcal{P}}\left[H_i\log\left(\pi\left(\boldsymbol{x}_i\right)\right)+\left(1-H_i\right)\log\left(1-\pi\left(\boldsymbol{x}_i\right)\right)\right]\nonumber\\
&+\sum_{i\in\mathcal{P}}\left[H_i\log\left(f_1\left(p_i\mid\boldsymbol{x}_i\right)\right)\right],
\end{align}
where we use the fact $\log\left(f_0\left(p_i\mid\boldsymbol{x}_i\right)\right)=\log\left(1\right)=0$. Because all $H_i$'s and parts of $p_i$'s are not observed, the Expectation-Maximization (EM) algorithm is an iterative algorithm to maximize the observed log-likelihood. For more information about missing data, see Chapter 3 in \citep{kim2021statistical}. We use $\mathcal{T}_t$ to denote the index set, $\mathcal{T}_t:=\{i\in\mathcal{P}:s_t\left(\boldsymbol{x}_i\right)<p_i<1-s_t\left(\boldsymbol{x}_i\right)\}$. The $p_i$ is known for $i\in\mathcal{T}_t$ at the $t$-th iteration. The detailed procedure is shown in Algorithm \ref{al:em}.
\begin{algorithm}
		\caption{The EM Algorithm}
		\label{al:em}
		\begin{algorithmic}
			\Require data $\{\boldsymbol{x}_i,\tilde{p}_{pm,t,i}\}_{i\in\mathcal{P}}$, initial value $\{\hat{\pi}^{\left(0\right)}\left(\cdot\right),\hat{f}_1^{\left(0\right)}\left(\cdot\mid\boldsymbol{x}\right)\}$, number of iteration $k$;
			\ForAll  {$r=1,\dots,k-1$}
			\State  [E-step]: Calculate $\mathbb{E}\left[l_t\left(\hat{\pi}^{\left(r\right)}\left(\cdot\right),\hat{f}^{\left(r\right)}_1\left(\cdot\mid\boldsymbol{x}\right)\right)\mid\{\boldsymbol{x}_i,\tilde{p}_{pm,t,i}\}_{i\in\mathcal{P}}\right]$;
                \State For $i\in\mathcal{P}$ and $i\in\mathcal{T}_t$:
                \State \quad $\hat{H}_i^{\left(r\right)}=\frac{\pi^{\left(r\right)}\left(\boldsymbol{x}_i\right)\hat{f}_1^{\left(r\right)}\left(p_i\mid\boldsymbol{x}_i\right)}{\pi^{\left(r\right)}\left(\boldsymbol{x}_i\right)\hat{f}_1^{\left(r\right)}\left(p_i\mid\boldsymbol{x}_i\right)+1-\pi^{\left(r\right)}\left(\boldsymbol{x}_i\right)}$;
                \State \quad $\widehat{\log}\left(f_1\left(p_i\mid\boldsymbol{x}_i\right)\right)^{\left(r\right)}=\log\left(f_1\left(p_i\mid\boldsymbol{x}_i\right)\right)$;
                \State For $i\in\mathcal{P}$ and $i\notin\mathcal{T}_t$:
                \State \quad $\hat{H}_i^{\left(r\right)}=\frac{\pi^{\left(r\right)}\left(\boldsymbol{x}_i\right)\left[\hat{f}_1^{\left(r\right)}\left(p_i\mid\boldsymbol{x}_i\right)+\hat{f}_1^{\left(r\right)}\left(1-p_i\mid\boldsymbol{x}_i\right)\right]}{\pi^{\left(r\right)}\left(\boldsymbol{x}_i\right)\left[\hat{f}_1^{\left(r\right)}\left(p_i\mid\boldsymbol{x}_i\right)+\hat{f}_1^{\left(r\right)}\left(1-p_i\mid\boldsymbol{x}_i\right)\right]+2\left[1-\pi^{\left(r\right)}\left(\boldsymbol{x}_i\right)\right]}$;
                \State \quad $\widehat{\log}\left(f_1\left(p_i\mid\boldsymbol{x}_i\right)\right)^{\left(r\right)}=\frac{\hat{f}_1^{\left(r\right)}\left(p_i\mid\boldsymbol{x}_i\right)\log\left(f_1\left(p_i\mid\boldsymbol{x}_i\right)\right)+\hat{f}_1^{\left(r\right)}\left(1-p_i\mid\boldsymbol{x}_i\right)\log\left(f_1\left(1-p_i\mid\boldsymbol{x}_i\right)\right)}{\hat{f}_1^{\left(r\right)}\left(p_i\mid\boldsymbol{x}_i\right)+\hat{f}_1^{\left(r\right)}\left(1-p_i\mid\boldsymbol{x}_i\right)}$;
                \State [M-step]: Solve 
                \begin{align*}
\{\hat{\pi}^{\left(r+1\right)}\left(\cdot\right),\hat{f}_1^{\left(r+1\right)}\left(\cdot\mid\boldsymbol{x}\right)\}=\arg\max& \mathbb{E}\left[l_t\left(\hat{\pi}^{\left(r\right)}\left(\cdot\right),\hat{f}^{\left(r\right)}_1\left(\cdot\mid\boldsymbol{x}\right)\right)\mid\{\boldsymbol{x}_i,\tilde{p}_{pm,t,i}\}_{i\in\mathcal{P}}\right]  \\
=\arg\max&\sum_{i\in\mathcal{P}}\left[\hat{H}_i^{\left(r\right)}\log\left(\pi\left(\boldsymbol{x}_i\right)\right)+\left(1-\hat{H}_i^{\left(r\right)}\right)\log\left(1-\pi\left(\boldsymbol{x}_i\right)\right)\right]\nonumber\\
&+\sum_{i\in\mathcal{P}}\left[\hat{H}_i^{\left(r\right)}\widehat{\log}\left(f_1\left(p_i\mid\boldsymbol{x}_i\right)\right)^{\left(r\right)}\right];
                \end{align*}
			\EndFor 
\Ensure $\{\hat{\pi}^{\left(k\right)}\left(\cdot\right),\hat{f}_1^{\left(k\right)}\left(\cdot\mid\boldsymbol{x}\right)\}$.
		\end{algorithmic}
	\end{algorithm}

At the $t$-th iteration with fitted model $\{\hat{\pi}^{\left(k\right)}\left(\cdot\right),\hat{f}_1^{\left(k\right)}\left(\cdot\mid\boldsymbol{x}\right)\}$, the second step is to select one hypothesis from $\mathcal{T}^c_t:=\{i\in\mathcal{P}:p_i\leq s_t\left(\boldsymbol{x}_i\right)\}$ and reject. The estimated probability of $H=0$ conditional on $\left(\boldsymbol{x},p\right)$ is
\begin{equation}
\label{eq:propa}
\mathbb{P}\left(H=0\mid\boldsymbol{x},p,\{\hat{\pi}^{\left(k\right)}\left(\cdot\right),\hat{f}_1^{\left(k\right)}\left(\cdot\mid\boldsymbol{x}\right)\}\right)=\frac{1-\pi^{\left(k\right)}\left(\boldsymbol{x}\right)}{\pi^{\left(k\right)}\left(\boldsymbol{x}\right)\hat{f}_1^{\left(k\right)}\left(p\mid\boldsymbol{x}\right)+1-\pi^{\left(k\right)}\left(\boldsymbol{x}\right)}.
\end{equation}
We propose to select the hypothesis with the largest probability defined in equation (\ref{eq:propa}) among the candidate set $\mathcal{T}_t^c$. Because all $p$-values in the candidate set are partially masked, we use the minimum elements in each pair and let $\tilde{p}_{pm,t,i}^{\prime}=\min\{\tilde{p}_i,1-\tilde{p}_i\}$ for $i\in\mathcal{T}_t^c$. As a consequence, we reject the $i$-th hypothesis for $i\in\mathcal{T}_t^c$ satisfying 
\begin{equation}
\label{eq:select}
i=\arg\max_{j\in\mathcal{T}_t^c}\mathbb{P}\left(H=0\mid\boldsymbol{x}_i,\tilde{p}_{pm,t,i}^{\prime},\{\hat{\pi}^{\left(k\right)}\left(\cdot\right),\hat{f}_1^{\left(k\right)}\left(\cdot\mid\boldsymbol{x}\right)\}\right).  
\end{equation}
We remark that the proposed selection criterion (\ref{eq:select}) is slightly different from the criterion in AdaPT procedure by \cite{lei2018adapt}. Under the conservative identifying assumption that
\begin{align*}
  1-\pi\left(\boldsymbol{x}\right)&=\inf_{p\in\left[0,1\right]}\left[\pi\left(\boldsymbol{x}\right)f_1\left(p\mid\boldsymbol{x}\right)+\left(1-\pi\left(\boldsymbol{x}\right)\right)f_0\left(p\mid\boldsymbol{x}\right)\right]\\
&=\pi\left(\boldsymbol{x}\right)f_1\left(1\mid\boldsymbol{x}\right)+\left(1-\pi\left(\boldsymbol{x}\right)\right)f_0\left(1\mid\boldsymbol{x}\right),
\end{align*}
the proposed selection criterion coincides with equation (23) in \citep{lei2018adapt}.

 \section{Numerical Illustrations}
	
	In this section, we numerically evaluate the performance of the proposed DP-AdaPT in terms of false discovery rate and power. We compare with three other methods: the original AdaPT without privacy guarantee \citep{lei2018adapt}, the differentially private Benjamini–Hochberg procedure (``DP-BH") proposed by \cite{dwork2021differentially}, and the private Bonferroni's method (``DP-Bonf") as discussed in \cite{dwork2021differentially}. 

	\subsection{Without Side Information}
	
	We first consider the case where only the $p$-values are obtained for each hypothesis and side information is unavailable. To ensure a fair comparison, we adopt the same simulation settings as in \cite{dwork2021differentially} and apply the noises with the same variance for all the differentially private methods. Specifically, we set the total number of hypotheses to be $n=\numprint{100000}$, with the number of true effects $t=100$. We select $m=500$ in the peeling step. Let $p_i=\Phi\left(\xi_i-\beta\right)$ for $i=1,\dots,t$, where $\Phi\left(\cdot\right)$ is the CDF of standard normal distribution and $\xi_1,\dots,\xi_m$ are i.i.d. standard normal distribution. We set the signal $\beta$ to be 4 and the significance level $\alpha=0.1$. Other parameters are required for the DP-BH algorithm, which is summarized in Algorithm \ref{alg:dp-bh}. Two parameters are used to control the sensitivity of the $p$-values in the DP-BH algorithm: the multiplicative sensitivity $\eta$ and the truncation threshold $\nu$. We set $\eta$ as $\numprint{0.0001}$ and $\nu=0.5\alpha/n$, which are the same as \cite{dwork2021differentially}. The privacy parameters are also set to be the same as in \cite{dwork2021differentially}: $\epsilon=0.5$ and $\delta=\numprint{0.001}$. For our proposed DP-AdaPT procedure, we set the privacy parameter $\mu=4\epsilon/\sqrt{10\log\left(1/\delta\right)}$. We use Gaussian CDF as the sensitivity transformation $G\left(\cdot\right)=\Phi\left(\cdot\right)$ and set the sensitivity parameter $\Delta_G=\eta$. The variance of the noise in our proposed DP-AdaPT procedure is the same as the variance of the noise in DP-BH.

	\begin{figure}[ht]
		\centering
		\centering
		\includegraphics[width=0.95\textwidth]{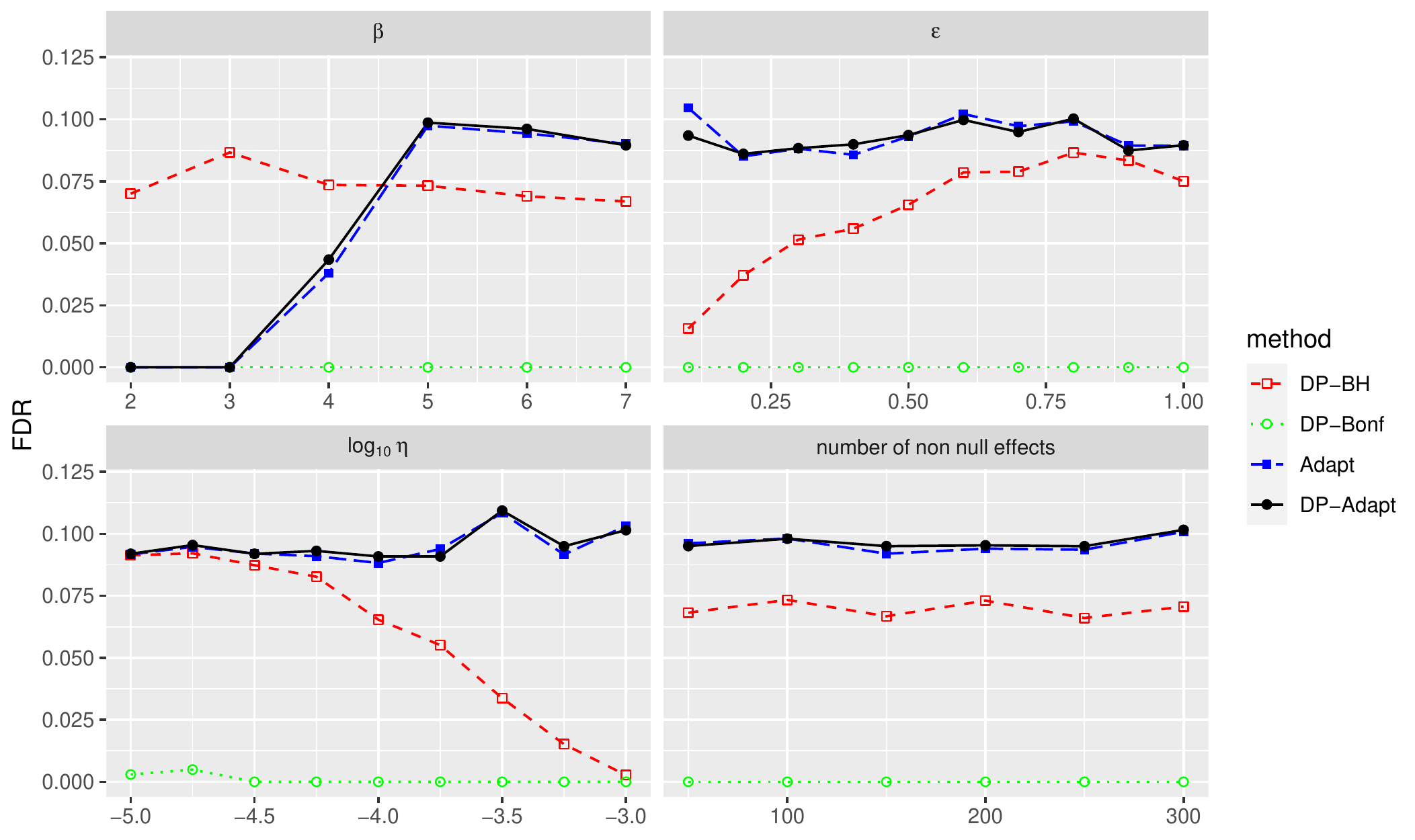}
		\caption{The FDR of DP-BH, DP-Bonf, and DP-AdaPT for varying parameters and averaged over $100$ independent trials.}
		\label{sim1_fdr}
	\end{figure}
	\begin{figure}[ht]
		\centering
		\includegraphics[width=0.95\textwidth]{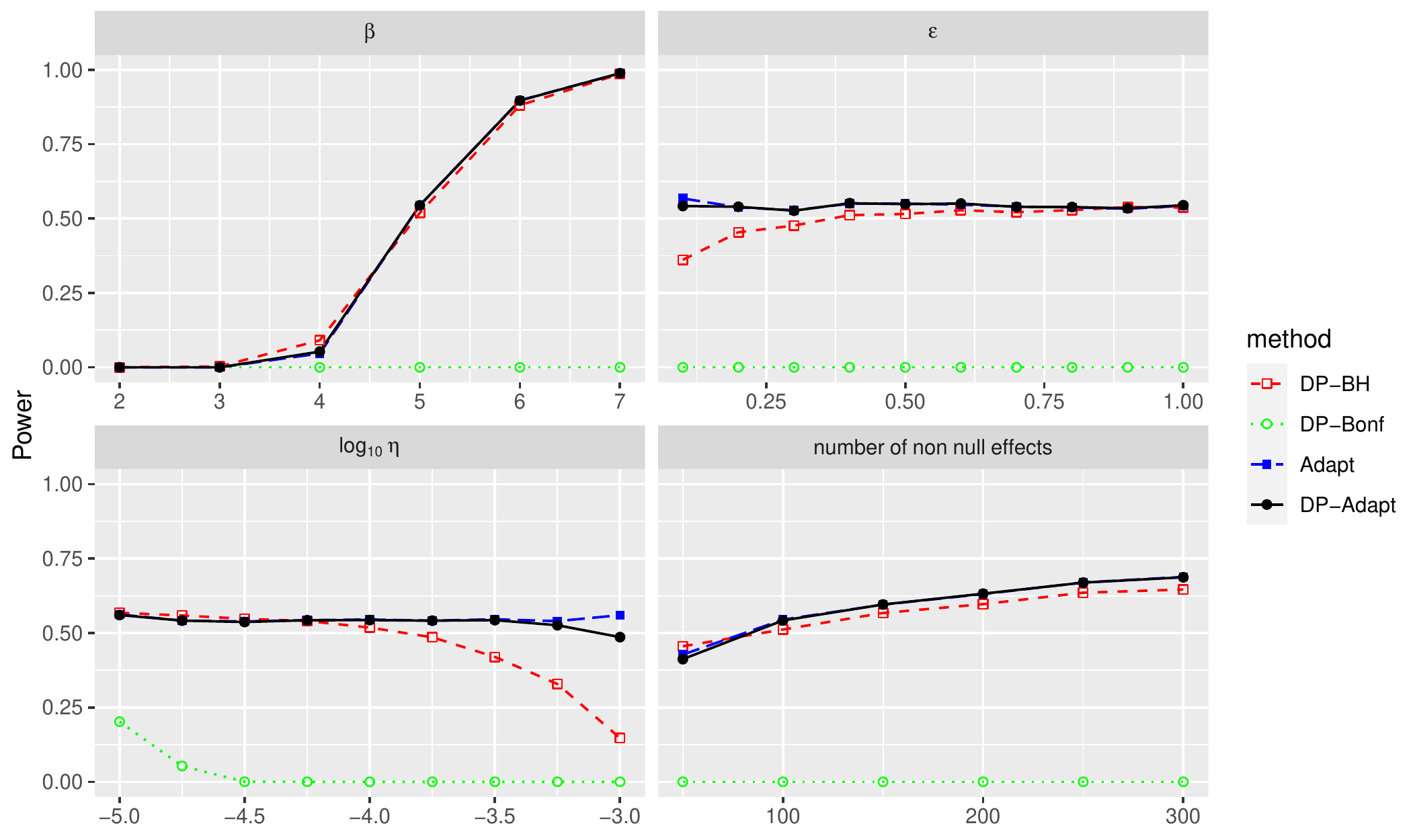}
		\caption{The power of DP-BH, DP-Bonf, and DP-AdaPT for varying parameters and averaged over $100$ independent trials.}
		\label{sim1_power}
	\end{figure}
	
	We first consider the situation where the null $p$-values all follow an independent uniform distribution. Specifically, we generate $p_i$ for $i=t+1,\dots,n$ independently from $U(0,1)$. 
	In Figure \ref{sim1_fdr}, we report the empirical FDR control for all the methods by varying the signal parameter $\beta$, the privacy parameter $\epsilon$, the sensitivity parameter $\eta$ and the number of true effects $t$. Clearly, all methods successfully control the FDR below the specified level $\alpha=0.1$. We report the power of all the methods in Figure \ref{sim1_power}. The naive DP-Bonf method is too conservative in detecting any positive signals.
	The power of DP-BH and DP-AdaPT performs similarly to each other in most cases. When the sensitivity parameter $\eta$ is large, the proposed method has better power than the DP-BH procedure. The rationale is that when $\eta$ is large, the variance of the noise is large, and the correction term in Algorithm \ref{alg:dp-bh}, i.e., $\eta\sqrt{10m\log\left(1/\delta\right)}\log\left(6m/\alpha\right)/\epsilon$, is large. The correction term plays the role of ruling out the influence of adding noise and guarantees FDR control with high probability. This is the main weakness of the DP-BH procedure. On the other hand, the proposed DP-AdaPT is based on the symmetry of $p$-values and provides valid finite sample FDR control, which is more robust both theoretically and practically.
	
\begin{figure}[ht]
		\centering
		\includegraphics[width=0.95\textwidth]{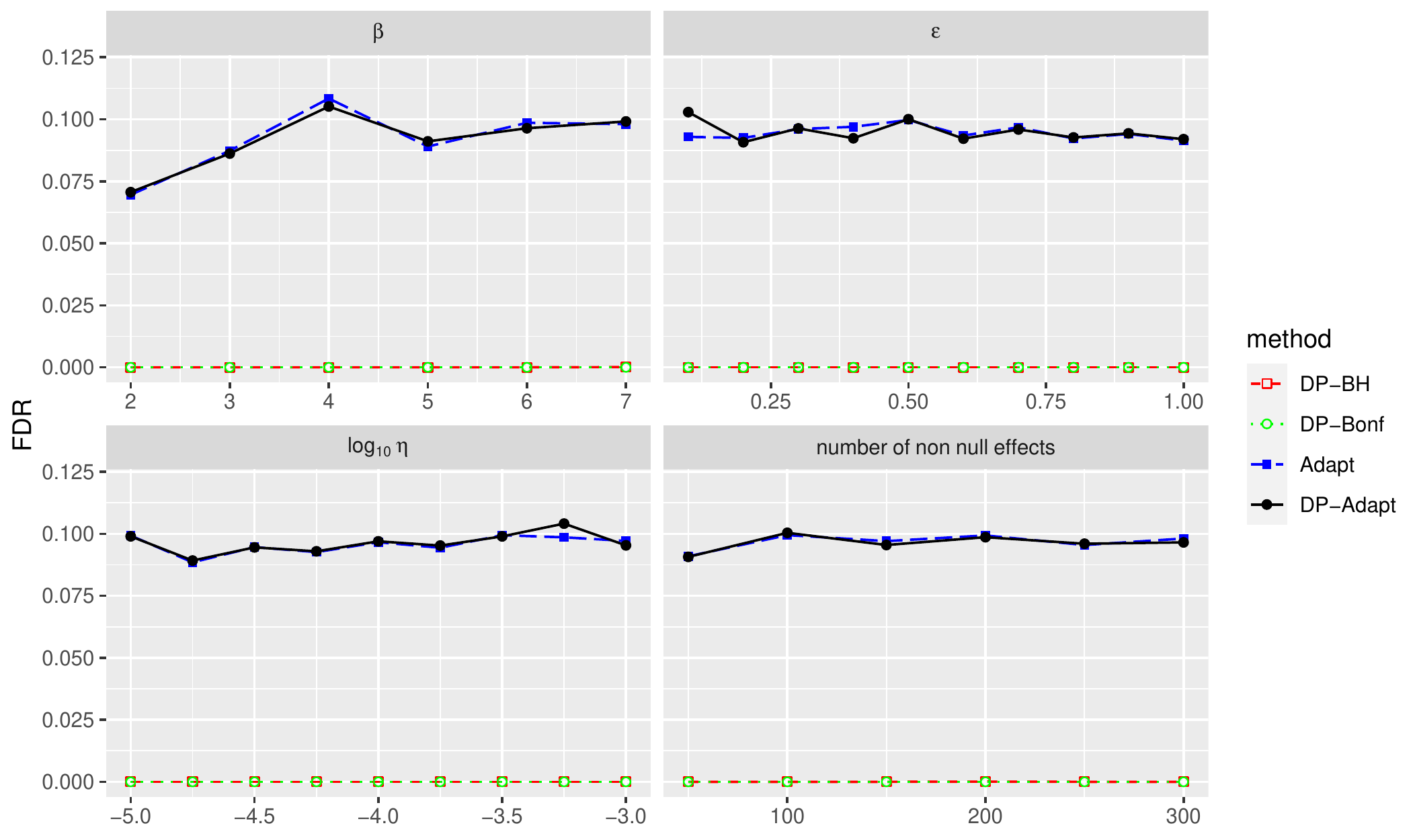}
		\caption{The FDR of DP-BH, DP-Bonf, and DP-AdaPT for varying parameters and averaged over $100$ independent trials under conservative $p$-values.}
		\label{sim1_fdr_con}
	\end{figure}

\begin{figure}[ht]
		\centering
		\includegraphics[width=0.95\textwidth]{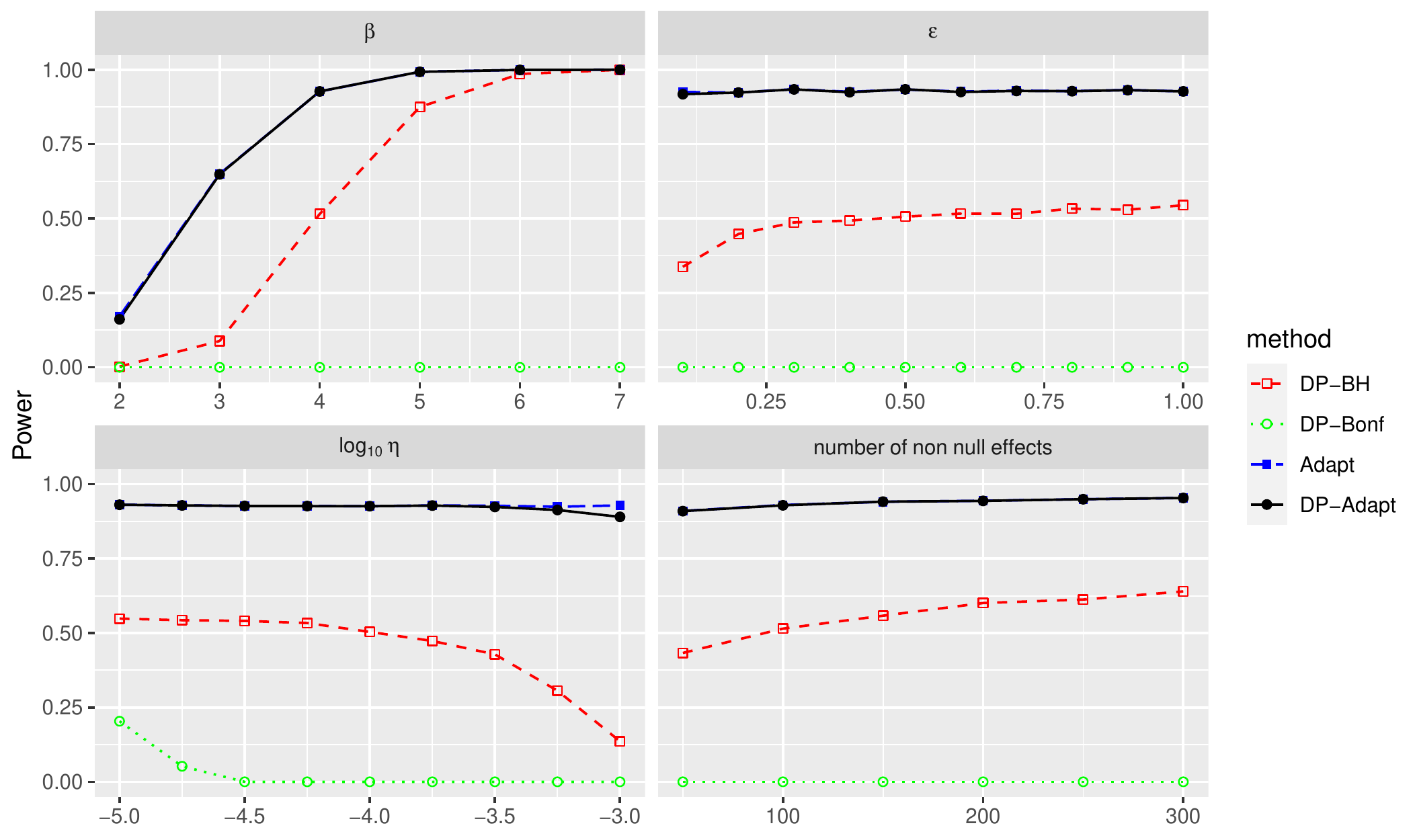}
		\caption{The power of DP-BH, DP-Bonf, and DP-AdaPT for varying parameters and averaged over $100$ independent trials under conservative $p$-values.}
		\label{sim1_power_con}
	\end{figure}

	Next, we consider the situation where the null $p$-values follow conservative distributions compared to the uniform, which is a common phenomenon in practice. Specifically, we generate $p_i$ for $i=t+1,\dots,n$ independently from Beta distribution with shape parameters $\left(2,2\right)$. The empirical FDR and power of all the methods are summarized in Figure \ref{sim1_fdr_con} and Figure \ref{sim1_power_con}. All methods successfully control the FDR below the $\alpha=0.1$. DP-BH and DP-Bonf have nearly zero false discovery rates. Though \cite{dwork2021differentially} only have a theoretical proof for FDR control when the $p$-value of null hypotheses follows the uniform distribution. It is not surprising that the DP-BH procedure controls FDR at the target level because the false discovery rates of conservative null hypotheses are easier to control than non-conservative null hypotheses in principle. The power of our proposed method is uniformly better than the DP-BH procedure. In general, our proposed method has power close to $0.90$ when the number of true effects is smaller than the number of invocations and the signal size is reasonably strong. 
	
	\subsection{With Side Information}
	
	In this subsection, we consider the case where the auxiliary side information is available for the hypothesis. We use similar simulation settings as in \cite{lei2018adapt}. The auxiliary covariates $\left(x_{1i},x_{2i}\right)$'s are generated from an equispaced $100\times 100$ grid in the area $\left[-100,100\right]\times\left[-100,100\right]$. The $p$-values are generated i.i.d. from $p_i=1-\Phi\left(z_i\right)$, where $z_i\sim N\left(\mu_i,1\right)$ and $\Phi\left(\cdot\right)$ is the CDF of $N\left(0,1\right)$. For $i\in\mathcal{H}_0$, we set $\beta_i=0$, and for $i\in\mathcal{H}_1$, we set $\beta_i=\beta$ for $\beta>0$. Three different patterns of $i\in\mathcal{H}_1$ are considered.
	$$i\in\mathcal{H}_1\Leftrightarrow \begin{cases}
		x_{1i}^2+x_{2i}^{2}\leq 150&\quad\text{\Rmnum{1}}\\
		\left(x_{1i}-65\right)^2+\left(x_{2i}-65\right)^{2}\leq 150&\quad\text{\Rmnum{2}} \\
		2\left(x_{1i}+x_{2i}\right)^2/100^2+\left(x_{2i}-x_{1i}\right)^2/15^2\leq 0.1&\quad\text{\Rmnum{3}} \\
	\end{cases}$$
	
	The number of the true signals $\mathcal{H}_1$ are $120,116$ and $118$ for case 1, case 2, and case 3, respectively. The number of selections in the peeling algorithm is set to $m=500$. For our proposed DP-AdaPT procedure, we use Gaussian CDF as the sensitivity transformation $G\left(\cdot\right)=\Phi\left(\cdot\right)$ and set the sensitivity parameter $\Delta_G=0.0001$. We set the privacy parameter $\mu=0.24$, which matches the noise scale in \cite{dwork2021differentially}. We compare the DP-AdaPT with the DP-BH method to evaluate the advantage of using side information. We also compare our proposed DP-AdaPT with the non-private AdaPT procedure to examine the privacy and accuracy tradeoff. For the AdaPT procedure, we follow a similar algorithm as in \cite{lei2018adapt} and fit two-dimensional Generalized Additive Models in M-step, using R package \textit{mgcv} with the knots selected automatically in every step by GCV criterion. The procedure replicates $100$ times, and the results are reported in terms of the average of $100$ replications.
	
\begin{figure}[ht]
		\centering
		\includegraphics[width=0.85\textwidth]{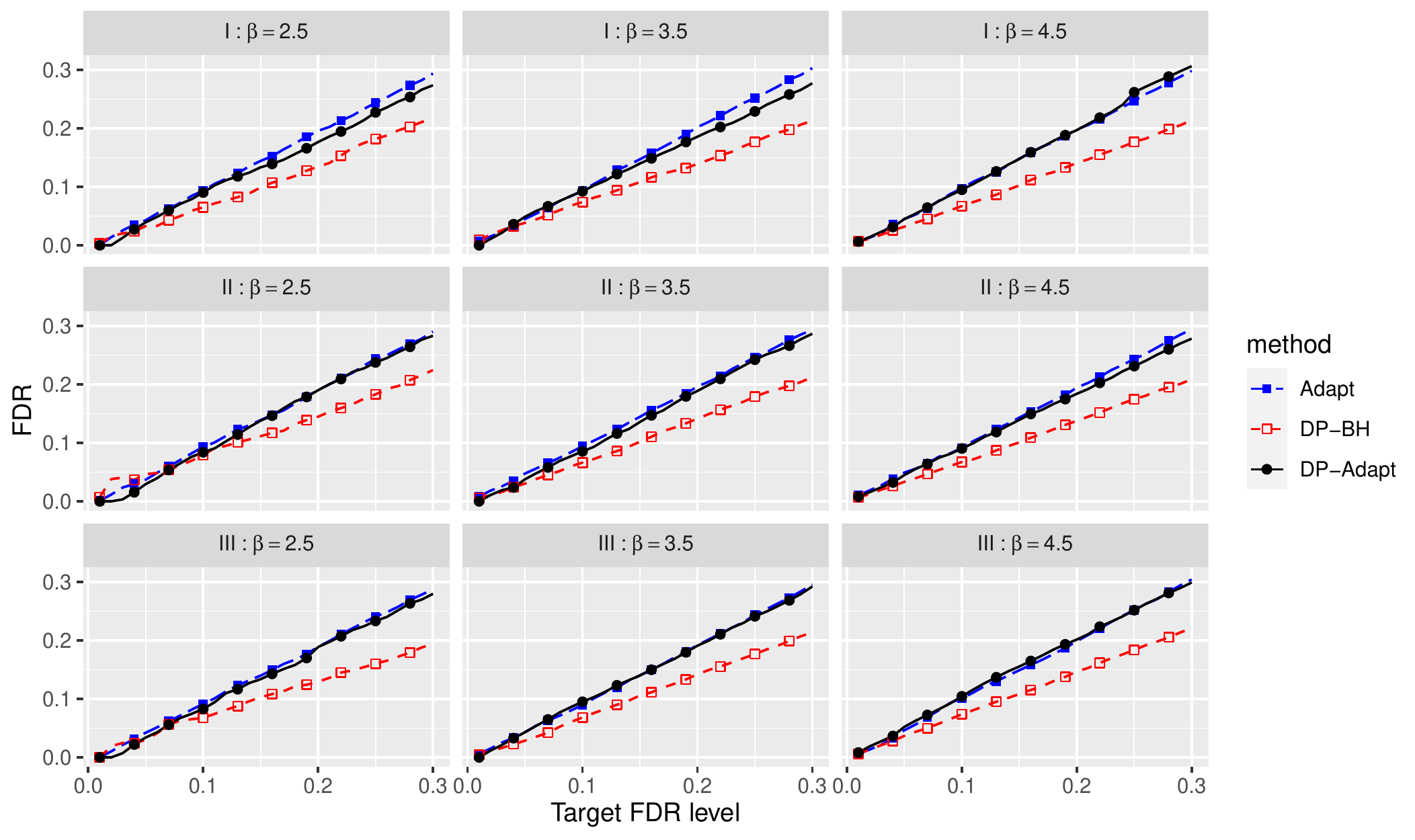}
		\caption{The FDR of DP-BH, AdaPT, and DP-AdaPT for varying signal strength and averaged over $100$ independent trials under uniform $p$-values.}
		\label{sim2_lei_fdr}
	\end{figure}

\begin{figure}[ht]
		\centering
		\includegraphics[width=0.85\textwidth]{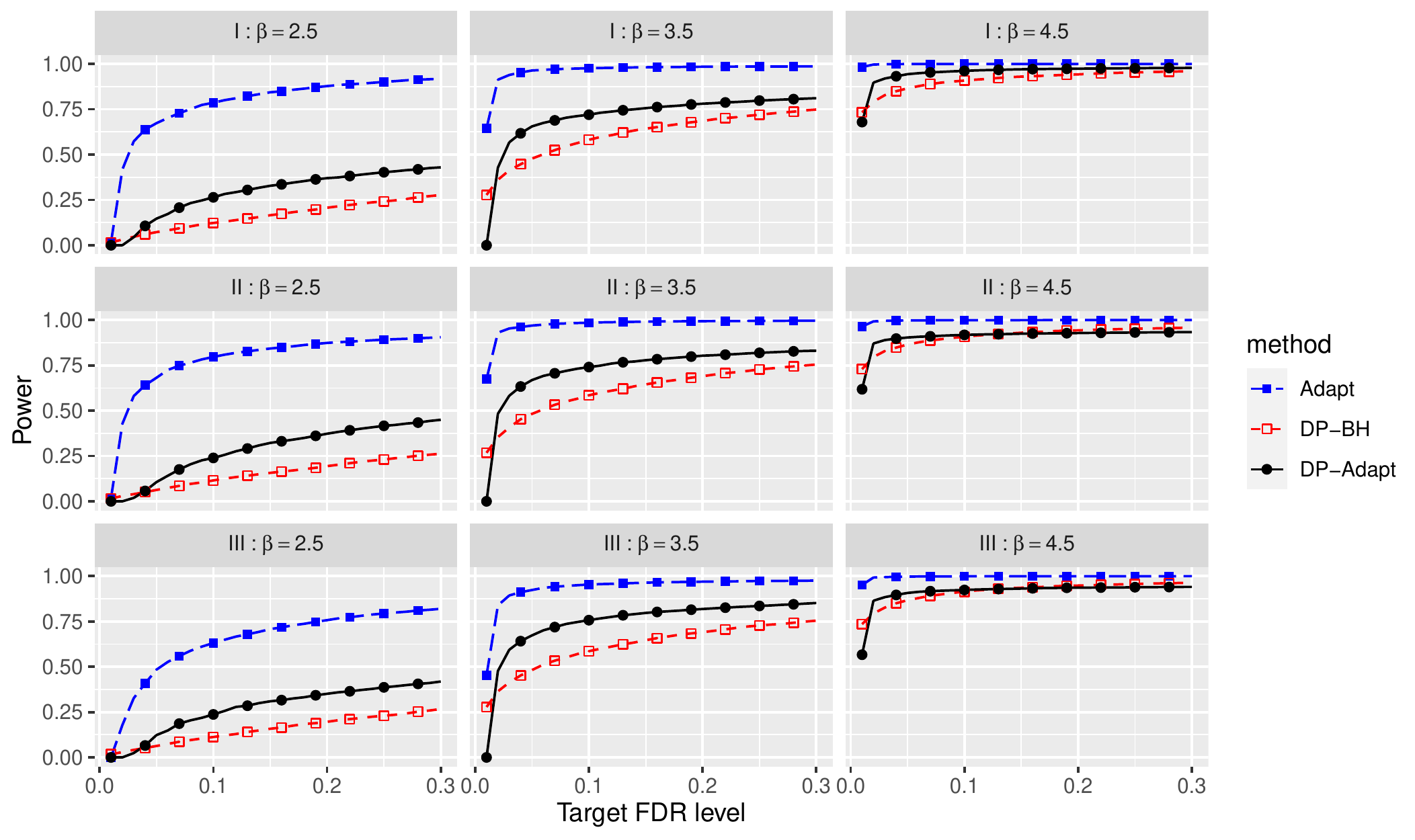}
		\caption{The power of DP-BH, AdaPT and DP-AdaPT for varying signal strength and averaged over $100$ independent trials under uniform $p$-values.}
		\label{sim2_lei_power}
	\end{figure}

	Figure \ref{sim2_lei_fdr} and Figure \ref{sim2_lei_power} present the empirical FDR and power of AdaPT, DP-BH, and DP-AdaPT, respectively. All methods can control the FDR at the desired level. The proposed DP-AdaPT procedure has larger power than the DP-BH procedure when the target FDR is greater than $0.03$. Especially when the strength of signals is not very strong, our proposed method has more than $50\%$ larger power than the DP-BH procedure. When the strength of signals is large enough, all procedures have similar powers, and AdaPT is slightly better than others. Compared to the original AdaPT, the proposed DP-AdaPT has uniformly smaller power, which is due to noise added for privacy guarantee. To be precise, the proposed procedure contains two pre-processing steps, peeling and adding noise. In the peeling algorithm, a subset of the original hypothesis is selected for further consideration. The selection procedure has two intrinsic weaknesses. First, some important variables can be potentially ignored due to random errors. Second, only $5\%$ data are used in the model fitting procedure of the DP-AdaPT. In the simulation settings, the side information is perfect for separating null hypotheses and alternative hypotheses. Adding noise to the $p$-values after selection attenuates the strength of a valid signal, which bring difficulties in rejecting the null hypothesis. To sum up, it is not surprising that DP-AdaPT is less efficient than AdaPT. However, the loss is relatively mild and is the cost of privacy.
	
\begin{figure}[ht]
		\centering
		\includegraphics[width=0.85\textwidth]{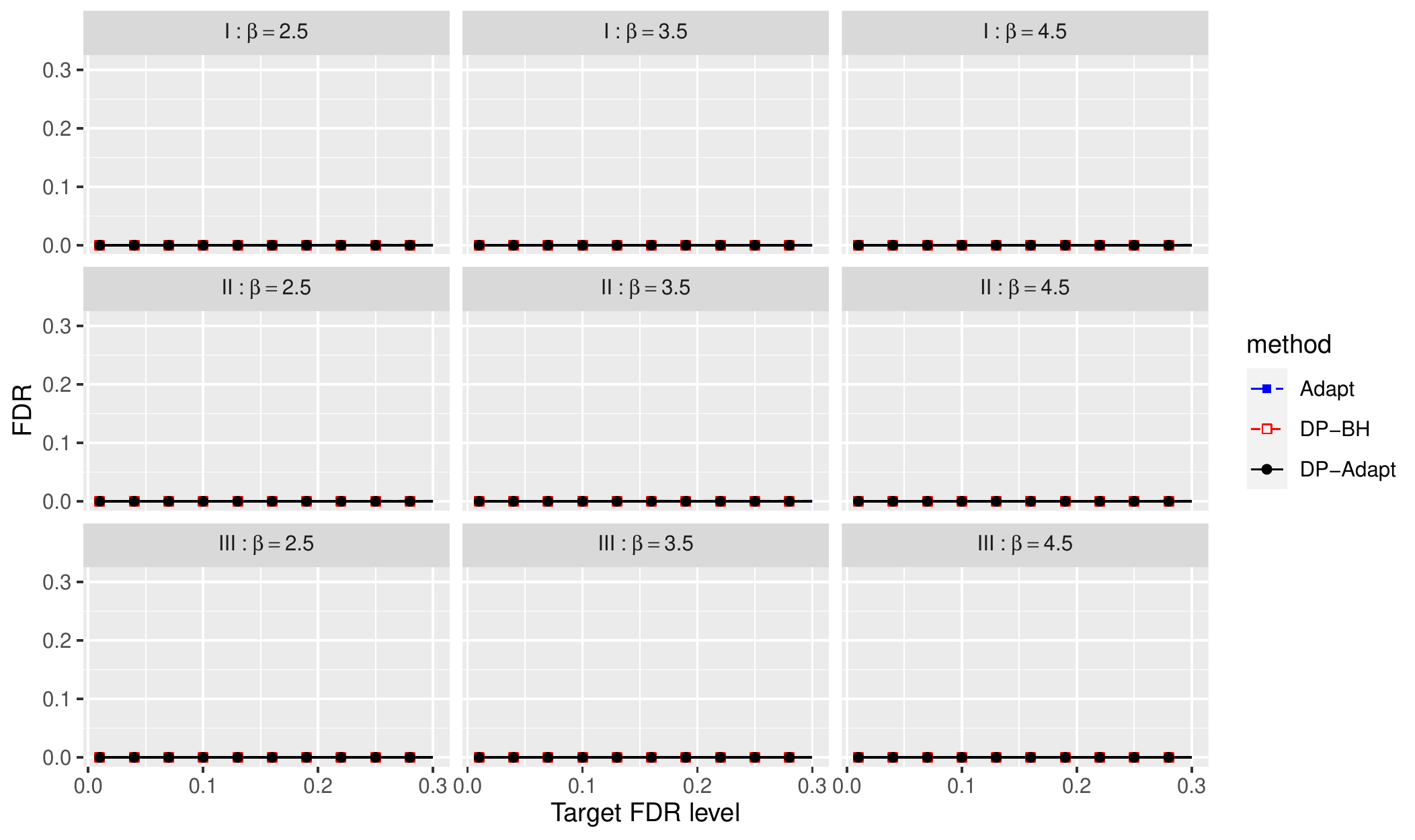}
		\caption{The FDR of DP-BH, AdaPT, and DP-AdaPT for varying signal strength and averaged over $100$ independent trials under conservative $p$-values.}
		\label{sim2_lei_con_fdr}
	\end{figure}

\begin{figure}[ht]
		\centering
		\includegraphics[width=0.85\textwidth]{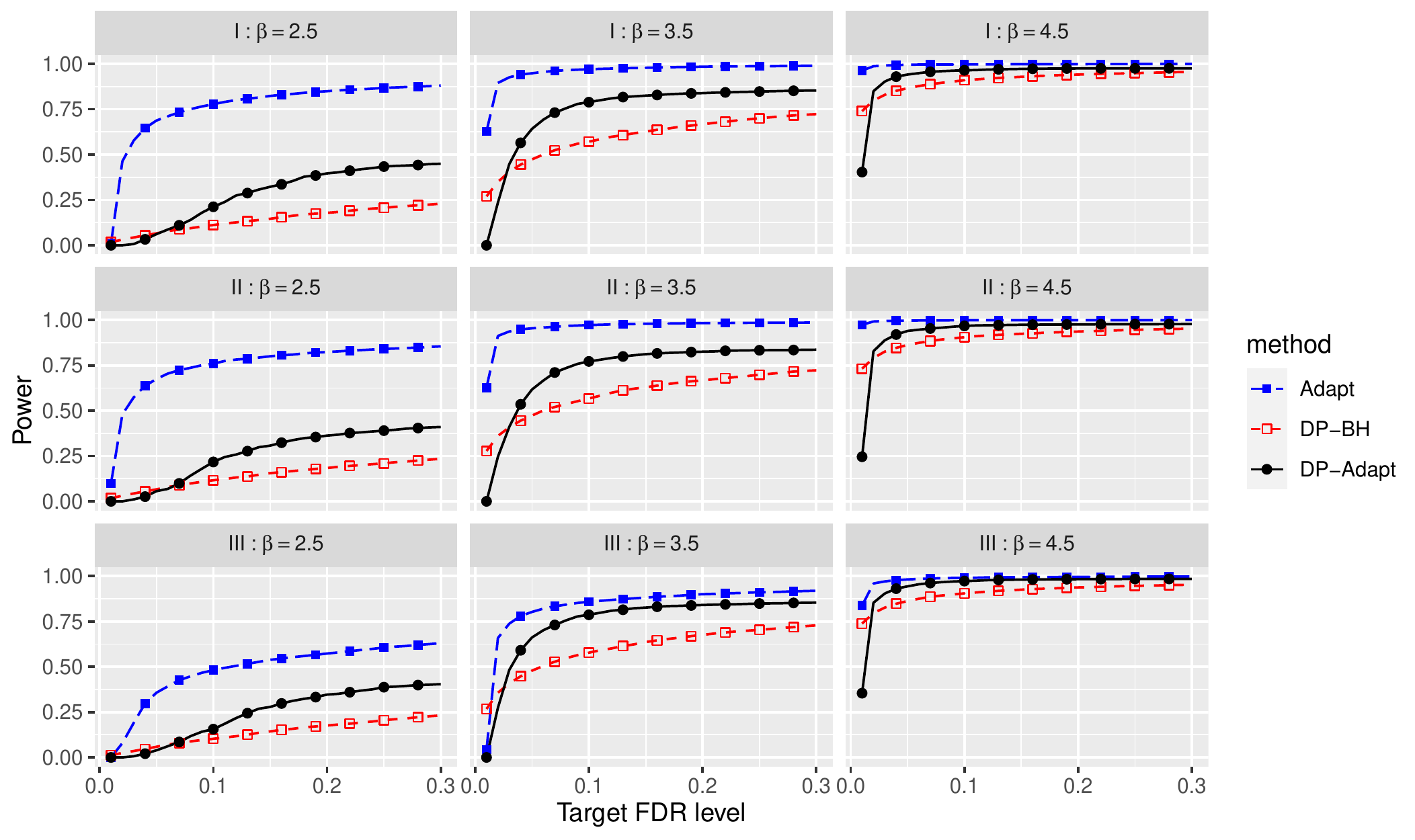}
		\caption{The power of DP-BH, AdaPT, and DP-AdaPT for varying signal strength and averaged over $100$ independent trials under conservative $p$-values.}
		\label{sim2_lei_con_power}
	\end{figure}

	We also consider the case where the null $p$-values are conservative. We modify the previous settings and generate $p_i$ for $i\in\mathcal{H}_0$ from density function $f\left(p\right)=4p^3$, $p\in\left[0,1\right]$. Figure \ref{sim2_lei_con_fdr} and Figure \ref{sim2_lei_con_power} present the empirical FDR and power of AdaPT, DP-BH, and DP-AdaPT, respectively. The distribution of null $p$-values is very conservative, and thus all methods successfully control the FDR. The proposed DP-AdaPT procedure has larger power than the DP-BH procedure when the target FDR is greater than $0.05$. Especially when the strength of the signal is not very strong, our proposed method is around $70\%$ better than the DP-BH procedure. Compared to the AdaPT procedure, the proposed DP-AdaPT has uniformly smaller power. When the strength of signals is reasonably strong, the difference between the power of our proposed DP-AdaPT and the power of AdaPT is smaller when the null is conservative. The power of the AdaPT procedure in the conservative case is smaller than that in the uniform case. Because the ratio of the density of $p$-values under the null hypothesis near $1$ and $0$ is very large when the density function is $f\left(p\right)=4p^3$, and thus the AdaPT is too conservative. However, the DP-AdaPT procedure uses noise to stable the ratio of the density of noisy $p$-values under the null hypothesis near $1$ and $0$, as illustrated in Figure \ref{fig:noisy_mirror}. Thus, our proposed procedure is more robust to the distribution of $p$-values.
	
	\newpage
	
	\begin{table}[ht]
		\centering
		\begin{tabular}{|r|r r|r r|r r|}
			\hline
			&\multicolumn{2}{c}{$\beta=2.5$}&\multicolumn{2}{|c|}{$\beta=3.5$}&\multicolumn{2}{c|}{$\beta=4.5$}\\
			\hline
			Uniform Null& AdaPT & DP-AdaPT & AdaPT & DP-AdaPT & AdaPT & DP-AdaPT \\ 
			\hline
			\Rmnum{1} & 1254.47 & 18.22 & 1100.75 & 20.06 & 1043.71 & 23.50 \\ 
			\Rmnum{2} & 1512.54 & 23.82 & 1314.59 & 23.01 & 1215.09 & 25.58 \\ 
			\Rmnum{3} & 1515.57 & 25.42 & 1090.52 & 16.74 & 1165.79 & 22.62 \\ 
			\hline
			Conservative Null& AdaPT & DP-AdaPT & AdaPT & DP-AdaPT & AdaPT & DP-AdaPT \\ 
			\hline
			\Rmnum{1} & 1810.04 & 30.34 & 1638.76 & 28.60 & 1662.77 & 26.49 \\ 
			\Rmnum{2} & 1798.20 & 26.85 & 1508.15 & 32.04 & 1770.05 & 29.97 \\ 
			\Rmnum{3} & 1921.82 & 31.59 & 1717.81 & 31.23 & 1675.85 & 25.46 \\ 
			\hline
		\end{tabular}
		\caption{Computation Time in seconds averaged over $100$ replications}
		\label{table:computation_time}
	\end{table}
	
	Although DP-AdaPT is not as accurate as the original AdaPT, the computation time of the proposed DP-AdaPT is much faster due to the mirror peeling algorithm. The model updating in AdaPT is computationally costly and should be performed every several steps. Thus when the dataset is large, the computation of AdaPT can be very slow. We provide the computation time in Table \ref{table:computation_time}. The computation time is based on an HPC cluster with CPU Model: Intel Xeon Gold 6152 and RAM: 10 GB. From the table, the computation of DP-AdaPT is much faster than AdaPT because the data size of DP-AdaPT is only $5\%$ of the data size of AdaPT. This provides a way to accelerate the computation of AdaPT in practice: when applying the AdaPT procedure to massive datasets, the proposed algorithm can be utilized to reduce the computation cost by applying noiseless screening of masked $p$-values.

\subsection{Empirical Example}

The Bottomly data set is an RNA-seq data set collected by \cite{bottomly2011evaluating} to detect differential striatal gene expression between the C57BL/6J (B6) and DBA/2J (D2) inbred mouse strains. An average of 22 million short sequencing reads were generated per sample for 21 samples (10 B6 and 11 D2). The Bottomly data was analyzed by \cite{ignatiadis2016data} using \textit{DESeq2} package \citep{love2014moderated} and performing an independent hypothesis weighting (IHW) method. Approximate Wald test statistics were used to calculate $p$-values. The logarithm of each gene's average normalized counts (across samples) is used as auxiliary information \citep{lei2018adapt}. After removing missing records, the data set contains $n=13 932$ genes ($p$-values).

\begin{figure}[ht]
		\centering
		\includegraphics[width=0.90\textwidth]{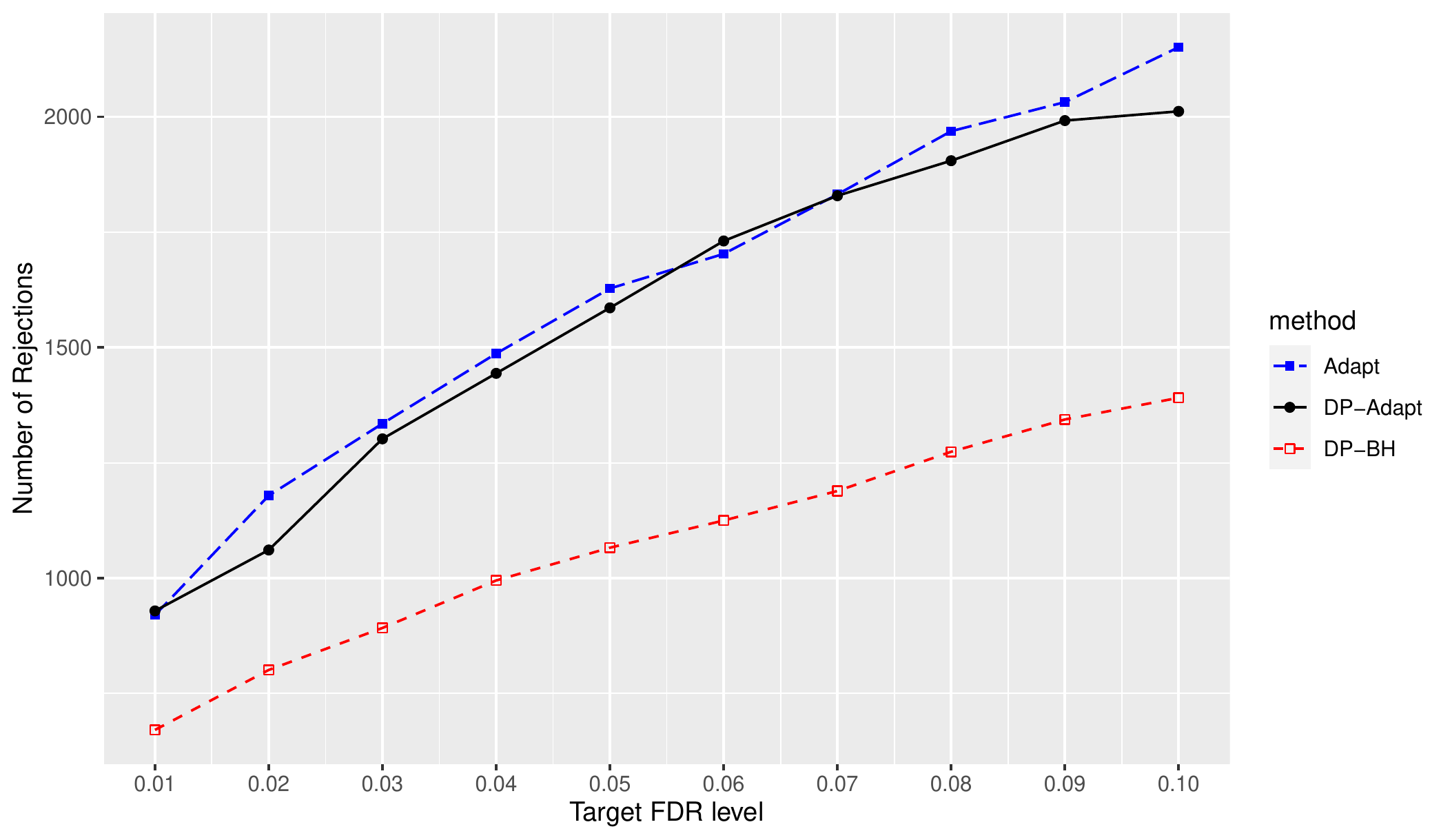}
		\caption{Bottomly dataset: number of rejections of DP-BH, AdaPT, and DP-AdaPT for varying FDR}
		\label{bottomly}
	\end{figure}

We pre-processed the data set by the \textit{DESeq2} package and used our proposed DP-AdaPT to identify differentially expressed genes while controlling the FDR at a pre-set level $\alpha=0.01,\dots,0.1$. Because $p$-values are calculated based on Wald statistics, we used the standard normal CDF $G\left(\cdot\right)=\Phi\left(\cdot\right)$ as illustrated in Example \ref{exam:one-side}. The privacy parameter was $\mu = 0.25$. The same privacy parameter was used by \cite{avella2021differentially}. The sensitive parameter is $\Delta=3\times10^{-5}$, which is roughly the inverse of the square root of the total sample size. The corresponding standard deviation of the added noise is $\sqrt{2}\sqrt{8m\Delta^2/\mu^2}\approx 0.024$. The number of pre-selected hypotheses in the peeling algorithm was $m=2500$, which is $18\%$ of the total number of hypotheses. We compare our proposed DP-AdaPT method with the non-private Adapt and private DP-BH methods. The results are shown in Figure \ref{bottomly}. 

Our proposed DP-AdaPT and Adapt methods have significantly more discoveries than the DP-BH method because the auxiliary variable is highly correlated with hypotheses. Our proposed DP-AdaPT procedure complies with the Adapt procedure, which matches the illustration in the simulation.

	\section{Discussion}
 In this paper, we propose a differentially private FDR control algorithm that accurately controls the FDR at a user-specified level with privacy guarantee. The proposed algorithm is based on adding noise to a transformed $p$-value, and preserves the mirror conservative property of the $p$-values. By further integrating a mirror peeling algorithm, we can define a nice filtration and apply the classic optimal stopping argument. Our analysis provides a new perspective in analyzing differentially private statistical algorithms: instead of conservatively controlling the noise added for privacy, researchers can design transformations or modifications that can preserve the key desired property in the theoretical analysis. 
	
	There are several open problems left for future research. For example, one commonly assumed conditions on the null $p$-values in multiple testing is the {\it uniformly conservative} property: $\mathbb{P}(p_i/\tau\leq t \mid p_i\leq\tau)\leq t$, $\forall\: t,\tau\in(0, 1)$. However, it is still unknown whether a similar transformation can be applied. If given a positive answer, then many existing multiple testing procedures can be easily extended to the differentially private version. It is an open question whether {\it mirror conservative} or {\it uniformly conservative} assumption is necessary for differentially private FDR control. The other commonly assumed conditions is the \textit{super uniform} property: $\mathbb{P}(p_i\leq t)\leq t$, $\forall\: t\in[0,1]$. Developing a differentially private algorithm with finite sample FDR control under a \textit{super uniform} framework remains an open problem. Moreover, it is also interesting to further develop other differentially private modern statistical inference tools.

	\bibliographystyle{refstyle.bst}
	\bibliography{ref}

	\newpage
	
	\appendix
\section*{Appendix A.}
	\section{Algorithm of DP-BH}
	
	\begin{algorithm}
		\caption{The Private BHq procedure \citep{dwork2021differentially}}\label{alg:dp-bh}
		\begin{algorithmic}
			\Require dataset $\mathcal{S}$, $p$-values $p_1,\dots,p_n$, threshold $\nu>0$, multiplicative sensitivity $\eta$, significant level $\alpha$, privacy level $\left(\epsilon,\delta\right)$, number of invocations $m$ and Laplace noise scale $\lambda=\eta\sqrt{10m\log\left(1/\delta\right)}/\epsilon$
			\Ensure a set of up to $m$ rejected hypotheses
			\State Apply the transformation and denote $f_i=\log\max\{\nu,p_i\left(\mathcal{S}\right)\}$ for $i=1,\dots,n$.
			\State Perform the Peeling Mechanism and Report Noisy Min algorithm with Laplace noise with scale at $\lambda$. The results are denoted by $\bigg\{\left(i_1,\tilde{f}_{i_1}\right),\dots,\left(i_{m^{\prime}},\tilde{f}_{i_{m}}\right)\bigg\}$.
			\ForAll  {$j=m$ to $1$}
			\If {$\tilde{f}_{i_j}>\log\left(\alpha j/n\right)-\eta\sqrt{10m\log\left(1/\delta\right)}\log\left(6m/\alpha\right)/\epsilon$} \State continue
			\Else 
			\State reject $p_{i_1},\dots,p_{i_j}$ and halt
			\EndIf
			\EndFor
		\end{algorithmic}
	\end{algorithm}
	
	\section{Proof of Theorem \ref{lem: noisy_p_value_mc}}
	
	\begin{proof}
		Without loss of generality, we assume the bound $U=\infty$ and the measure $\mu$ is the Lebesgue measure. All other cases can be proved using the same technique by using a suitable measure.
		
		We first observe that the random variable $\tilde{p}$ is supported in $\left[0,1\right]$. For any $0\leq a_1\leq a_2\leq 0.5$, it remains to show the mirror-conservative condition, $$\mathbb{P}\left(\tilde{p}\in\left[a_1,a_2\right]\right)\leq \mathbb{P}\left(\tilde{p}\in\left[1-a_2,1-a_1\right]\right).$$
		By the definition, $\tilde{p}=G\{G^{-1}\left(p\right)+Z\}$,	the mirror-conservative condition is equivalent to
		$$\mathbb{P}\left(G^{-1}\left(p\right)+Z\in\left[G^{-1}\left(a_1\right),G^{-1}\left(a_2\right)\right]\right)\leq \mathbb{P}\left(G^{-1}\left(p\right)+Z\in\left[G^{-1}\left(1-a_2\right),G^{-1}\left(1-a_1\right)\right]\right),$$
		where we substitute $\tilde{p}$ in the equation and apply the function $G^{-1}\left(\cdot\right)$ to both sides.
		By the symmetry of $g\left(\cdot\right)$, we have for $x<0$, 
		$$G\left(x\right)=\int_{-\infty}^{x}g\left(t\right)dt=\int_{-\infty}^{x}g\left(-t\right)dt=\int_{-x}^{\infty}g\left(t\right)dt=1-\int_{-\infty}^{-x}g\left(t\right)dt=1-G\left(-x\right),$$
		where we use the fact $g\left(x\right)=g\left(-x\right)$ and $\int_{-\infty}^{\infty}g\left(x\right)=1$. After subtracting $1/2$ on both sides, we have $1/2-G\left(x\right)=-1/2+G\left(-x\right)$. Noticing that $G\left(x\right)<1/2$ for $x<0$, we have 
		$$G^{-1}\left(1/2-t\right)=-G^{-1}\left(1/2+t\right),$$
		where $t=G\left(x\right)\in\left(0,1/2\right)$ and we apply $G^{-1}\left(\cdot\right)$ to both sides. It follows that $G^{-1}\left(1-a_1\right)=-G^{-1}\left(a_1\right)$ and $G^{-1}\left(1-a_2\right)=-G^{-1}\left(a_2\right)$ by substituting $t$ by $1/2-a_1$ and $1/2-a_2$, respectively. 
		The mirror-conservative condition is equivalent to		$$\mathbb{P}\left(G^{-1}\left(p\right)+Z\in\left[G^{-1}\left(a_1\right),G^{-1}\left(a_2\right)\right]\right)\leq \mathbb{P}\left(G^{-1}\left(p\right)+Z\in\left[-G^{-1}\left(a_2\right),-G^{-1}\left(a_1\right)\right]\right),$$
		where we use the relation $G^{-1}\left(1-a_1\right)=-G^{-1}\left(a_1\right)$ and $G^{-1}\left(1-a_2\right)=-G^{-1}\left(a_2\right)$.
		
		To prove the mirror-conservative condition, we first perform a decomposition of the distribution of $p$ and use the convolution formula. Let $\mu_p$ be the corresponding probability measure generated by the distribution of $p$. We define a symmetric measure $\mu_s$ on $\left[0,1\right]$, which satisfies that $\mu_s\left(a,b\right]=\mu\left(a,b\right]$ for $0\leq a\leq b\leq 1/2$ and $\mu_s\left(a,b\right]=\mu\left(1-b,1-a\right]$ for $1/2\leq a\leq b\leq 1$. Carathéodory's extension theorem guarantees the existence of $\mu_s$. Intuitively, the $\mu_s$ is the mirror-symmetric part of $\mu_p$. We have the decomposition $\mu_p=\mu_s+\mu_r$, where $\mu_z$ is defined by $\mu_r:=\mu_p-\mu_s$. The $\mu_r$ is a positive measure by the fact that $\mu_p$ is mirror-conservative. Let $\mu_z$ be the probability measure generated by the standard normal distribution. It follows that
		\begin{align*}		&\mathbb{P}\left(G^{-1}\left(p\right)+Z\in\left[G^{-1}\left(a_1\right),G^{-1}\left(a_2\right)\right]\right)- \mathbb{P}\left(G^{-1}\left(p\right)+Z\in\left[-G^{-1}\left(a_2\right),-G^{-1}\left(a_1\right)\right]\right)\\
			&=\left(\mu_p\circ G\right)\otimes\mu_z\left[G^{-1}\left(a_1\right),G^{-1}\left(a_2\right)\right]-\left(\mu_p\circ G\right)\otimes\mu_z\left[-G^{-1}\left(a_2\right),-G^{-1}\left(a_1\right)\right]\\
			&=\left(\mu_s\circ G\right)\otimes\mu_z\left[G^{-1}\left(a_1\right),G^{-1}\left(a_2\right)\right]-\left(\mu_s\circ G\right)\otimes\mu_z\left[-G^{-1}\left(a_2\right),-G^{-1}\left(a_1\right)\right]\\
			&+\left(\mu_r\circ G\right)\otimes\mu_z\left[G^{-1}\left(a_1\right),G^{-1}\left(a_2\right)\right]-\left(\mu_r\circ G\right)\otimes\mu_z\left[-G^{-1}\left(a_2\right),-G^{-1}\left(a_1\right)\right],
		\end{align*}
		where $\mu_p\circ G$ is the probability measure generated by the random variable $G^{-1}\left(p\right)$ and $\otimes$ is the convolution operator. 
		
		We consider the first line in the above equation. For all $-\infty\leq a\leq b\leq 0$, using the fact that the two intervals $\left[G\left(a\right),G\left(b\right)\right]$ and $\left[G\left(-b\right),G\left(-a\right)\right]=\left[1-G\left(b\right),1-G\left(a\right)\right]$ are symmetric around $1/2$ and the measure $\mu_s$ is symmetric around $1/2$, we know $\mu_s\circ G\left[a,b\right]=\mu_s\circ G\left[-b,-a\right]$. Thus, the measure $\mu_s\circ G$ is symmetric around $0$. By the fact that $\mu_z$ is symmetric around $0$, the convolution $\left(\mu_s\circ G\right)\otimes\mu_z$ is also symmetric around $0$. Thus, we have 
		$$\left(\mu_s\circ G\right)\otimes\mu_z\left[G^{-1}\left(a_1\right),G^{-1}\left(a_2\right)\right]=\left(\mu_s\circ G\right)\otimes\mu_z\left[-G^{-1}\left(a_2\right),-G^{-1}\left(a_1\right)\right].$$
		
		It remains to consider the second line of the equation,
		$$\left(\mu_r\circ G\right)\otimes\mu_z\left[G^{-1}\left(a_1\right),G^{-1}\left(a_2\right)\right]-\left(\mu_r\circ G\right)\otimes\mu_z\left[-G^{-1}\left(a_2\right),-G^{-1}\left(a_1\right)\right].$$
		By the decomposition of $\mu_p$, we have $\mu_r\left(a,b\right]=0$ for $0\leq a\leq b\leq 1/2$. By the fact that $G\left(x\right)\leq 1/2$ for $x\leq 0$, we conclude that $\left(\mu_r\circ G\right)\left[a,b\right]=0$ for all $a,b\leq 0$. Let $\phi\left(\cdot\right)=d\mu_z/d\mu$ be the probability density function of standard normal distribution, and we have
		\begin{align*}
			&\left(\mu_r\circ G\right)\otimes\mu_z\left[G^{-1}\left(a_1\right),G^{-1}\left(a_2\right)\right]-\left(\mu_r\circ G\right)\otimes\mu_z\left[-G^{-1}\left(a_2\right),-G^{-1}\left(a_1\right)\right]\\
			&=\int_{-\infty}^{\infty}1\bigg\{x\in\left[G^{-1}\left(a_1\right),G^{-1}\left(a_2\right)\right]\bigg\}\int_{-\infty}^{\infty}\phi\left(x-y\right)d\left(\mu_r\circ G\right)\left(y\right)d\mu\left(x\right)\\
			&-\int_{-\infty}^{\infty}1\bigg\{x\in\left[-G^{-1}\left(a_2\right),-G^{-1}\left(a_1\right)\right]\bigg\}\int_{-\infty}^{\infty}\phi\left(x-y\right)d\left(\mu_r\circ G\right)\left(y\right)d\mu\left(x\right)\\
			&=\int_{G^{-1}\left(a_1\right)}^{G^{-1}\left(a_2\right)}\int_{0}^{\infty}\phi\left(x-y\right)d\left(\mu_r\circ G\right)\left(y\right)dx-\int_{-G^{-1}\left(a_2\right)}^{-G^{-1}\left(a_1\right)}\int_{0}^{\infty}\phi\left(x-y\right)d\left(\mu_r\circ G\right)\left(y\right)d\mu\left(x\right)\\
			&=\int_{G^{-1}\left(a_1\right)}^{G^{-1}\left(a_2\right)}\int_{0}^{\infty}\phi\left(x-y\right)d\left(\mu_r\circ G\right)\left(y\right)dx-\int_{G^{-1}\left(a_1\right)}^{G^{-1}\left(a_2\right)}\int_{0}^{\infty}\phi\left(-x-y\right)d\left(\mu_r\circ G\right)\left(y\right)d\mu\left(x\right)\\
			&=\int_{G^{-1}\left(a_1\right)}^{G^{-1}\left(a_2\right)}\int_{0}^{\infty}\left[\phi\left(x-y\right)-\phi\left(-x-y\right)\right]d\left(\mu_r\circ G\right)\left(y\right)d\mu\left(x\right)\\
			&=-\int_{G^{-1}\left(a_1\right)}^{G^{-1}\left(a_2\right)}\int_{0}^{-2x}\phi\left(-x-y\right)d\left(\mu_r\circ G\right)\left(y\right)d\mu\left(x\right).
		\end{align*}
		Remembering that $G^{-1}\left(a_1\right),G^{-1}\left(a_2\right)<0$, we have $x<0$. We conclude $$-\int_{G^{-1}\left(a_1\right)}^{G^{-1}\left(a_2\right)}\int_{0}^{-2x}\phi\left(-x-y\right)d\left(\mu_r\circ G\right)\left(y\right)dx<0.$$
		So, $$\mathbb{P}\left(\tilde{p}\in\left[a_1,a_2\right]\right)\leq \mathbb{P}\left(\tilde{p}\in\left[1-a_2,1-a_1\right]\right),$$ 
		and $\tilde{p}$ is mirror conservative.
	\end{proof}
	
	\section{Proof of Example \ref{exam:two-side} and \ref{exam:chi-sq}}
	
	\subsection{Example \ref{exam:two-side}}
	
	\begin{proof}
		For two-sided testing, we let the support $\text{supp}\{g\left(x\right)\}\subset \left[-M,M\right]$ and $g\left(M\right)=g\left(-M\right)>0$, and $G\left(x\right)=\int_{-\infty}^{x}g\left(x\right)dx$. We have
		\begin{align*}
			&G^{-1}\{2\Phi\left[-\left|T\left(\mathcal{S}^{\prime}\right)\right|\right]\}-G^{-1}\{2\Phi\left[-\left|T\left(\mathcal{S}\right)\right|\right]\}\\
			=&\int_{-\left|T\left(\mathcal{S}\right)\right|}^{-\left|T\left(\mathcal{S}^{\prime}\right)\right|}\frac{dG^{-1}\left[2\Phi\left(t\right)\right]}{dt}dt=\int_{-\left|T\left(\mathcal{S}\right)\right|}^{-\left|T\left(\mathcal{S}^{\prime}\right)\right|}\frac{dG^{-1}\left[2\Phi\left(t\right)\right]}{d 2\Phi\left(t\right)}\frac{d 2\Phi\left(t\right)}{dt}dt\\
			=&\int_{-\left|T\left(\mathcal{S}\right)\right|}^{-\left|T\left(\mathcal{S}^{\prime}\right)\right|}\frac{2\phi\left(t\right)}{g\left(G^{-1}\left(2\Phi\left(t\right)\right)\right)}dt.
		\end{align*}
		
		The function $2\phi\left(t\right)/g\left(G^{-1}\left(2\Phi\left(t\right)\right)\right)$ is continuous on $\left(-\infty,0\right)$. As $t\to -\infty$,
		\begin{align*}
			&\lim_{t\to -\infty}\frac{2\phi\left(t\right)}{g\left(G^{-1}\left(2\Phi\left(t\right)\right)\right)}=\sqrt{\frac{\pi}{2}}\lim_{t\to -\infty}\frac{\exp\left(-t^2/2\right)}{g\left(-M\right)}=0,
		\end{align*}
		where we used L'Hospital's rule. As $t\to 0^{-}$,
		\begin{align*}
			&\lim_{t\to 0^{-}}\frac{2\phi\left(t\right)}{g\left(G^{-1}\left(2\Phi\left(t\right)\right)\right)}=\lim_{t\to 0^{-}}\frac{2\phi\left(0\right)}{g\left(G^{-1}\left(2\Phi\left(t\right)\right)\right)}=\frac{2\phi\left(0\right)}{g\left(M\right)},
		\end{align*}
		where we used L'Hospital's rule again. Thus, the function is bounded by a constant $C$, and 
		\begin{align*}
			&\left|G^{-1}\{2\Phi\left[-\left|T\left(\mathcal{S}\right)\right|\right]\}-G^{-1}\{2\Phi\left[-\left|T\left(\mathcal{S}^{\prime}\right)\right|\right]\}\right|\leq\left|\int_{-\left|T\left(\mathcal{S}^{\prime}\right)\right|}^{-\left|T\left(\mathcal{S}\right)\right|}C\right|\\
			\leq & C\|\left|T\left(\mathcal{S}\right)\right|-\left|T\left(\mathcal{S}^{\prime}\right)\right|\|\leq C\|T\left(\mathcal{S}\right)-T\left(\mathcal{S}^{\prime}\right)\|\leq CM/\sqrt{n},
		\end{align*}
		where we use the fact $\sup\|T\left(\mathcal{S}^{\prime}\right)-T\left(\mathcal{S}\right)\|\leq M/\sqrt{n}$ by the sensitivity of statistic $T$.
	\end{proof}

	\subsection{Example  \ref{exam:chi-sq}}
	\begin{proof}
		We let $G\left(\cdot\right)=\Phi\left(\cdot\right)$, $\xi\left(x\right)=\left(2\pi\right)^{-1/2}x^{-1/2}e^{-x/2}$ be the density function of $\chi^2_1$ distribution and $\Xi$ be the CDF of $\chi^2_1$ distribution. We have
		\begin{align*}
			&\Phi^{-1}\{1-\Xi\left[T\left(\mathcal{S}\right)\right]\}-\Phi^{-1}\{1-\Xi\left[T\left(\mathcal{S}^{\prime}\right)\right]\}\\
			=&\int_{T\left(\mathcal{S}\right)}^{T\left(\mathcal{S}^{\prime}\right)}\frac{d\Phi^{-1}\left[1-\Xi\left(t\right)\right]}{dt}dt=\int_{T\left(\mathcal{S}\right)}^{T\left(\mathcal{S}^{\prime}\right)}\frac{d\Phi^{-1}\left[1-\Xi\left(t\right)\right]}{d\left[1-\Xi\left(t\right)\right]}\frac{d\left[1-\Xi\left(t\right)\right]}{dt}dt\\
			=&\int_{T\left(\mathcal{S}^{\prime}\right)}^{T\left(\mathcal{S}\right)}\frac{\xi\left(t\right)}{\phi\left(\Phi^{-1}\left(1-\Xi\left(t\right)\right)\right)}dt.
		\end{align*}
		The function $\xi\left(t\right)/\phi\left(\Phi^{-1}\left(1-\Xi\left(t\right)\right)\right)$ is continuous on $\left(0,\infty\right)$. We will use L'Hospital's rule to evaluate the limiting behavior of the function. The derivative of the denominator is,
		\begin{align*}
			&\frac{d\phi\left(\Phi^{-1}\left(1-\Xi\left(t\right)\right)\right)}{dt}=\frac{d\phi\left(\Phi^{-1}\left(1-\Xi\left(t\right)\right)\right)}{d\Phi^{-1}\left(1-\Xi\left(t\right)\right)}\frac{d\Phi^{-1}\left(1-\Xi\left(t\right)\right)}{d\left(1-\Xi\left(t\right)\right)}\frac{d\left(1-\Xi\left(t\right)\right)}{dt}\\
			=&-\Phi^{-1}\left(1-\Xi\left(t\right)\right)\phi\left(\Phi^{-1}\left(1-\Xi\left(t\right)\right)\right)\times\frac{1}{\phi\left(\Phi^{-1}\left(1-\Xi\left(t\right)\right)\right)}\times\left[-\xi\left(t\right)\right]\\
			=&\Phi^{-1}\left(1-\Xi\left(t\right)\right)\xi\left(t\right),
		\end{align*}
		where we use the fact 
		\begin{align*}
			\frac{d\phi\left(t\right)}{dt}=-t\frac{\exp\left(-t^2/2\right)}{\sqrt{2\pi}}=-t\phi\left(t\right).
		\end{align*}
		As $t\to \infty$,
		\begin{align*}
			&\lim_{t\to \infty}\frac{\xi\left(t\right)}{\phi\left(\Phi^{-1}\left(1-\Xi\left(t\right)\right)\right)}=\lim_{t\to \infty}\frac{1}{\sqrt{2\pi}}\frac{-\frac{1}{2}t^{-3/2}e^{-t/2}-\frac{1}{2}t^{-1/2}e^{-t/2}}{\Phi^{-1}\left(1-\Xi\left(t\right)\right)\xi\left(t\right)}\\
			=&\lim_{t\to \infty}-\frac{1}{2}\frac{\frac{1}{t}+1}{\Phi^{-1}\left(1-\Xi\left(t\right)\right)}\stackrel{u:=1-\Xi\left(t\right)}{=}\lim_{u\to 0^+}-\frac{1}{2}\frac{1}{\Phi^{-1}\left(u\right)}=0.
		\end{align*}
		As $t\to 0^+$, the function $\xi\left(t\right)/\phi\left(\Phi^{-1}\left(1-\Xi\left(t\right)\right)\right)$ diverges, which means we can not bound the sensitivity using the method in Example \ref{exam:two-side}. We investigate the limiting behavior of the function $\xi\left(t\right)/\phi\left(\Phi^{-1}\left(1-\Xi\left(t\right)\right)\right)$ by multiplying a power of $t$. For $0<\delta<1/2$,
		\begin{align*}
			&\lim_{t\to 0^{+}}t^{1/2+\delta}\frac{\xi\left(t\right)}{\phi\left(\Phi^{-1}\left(1-\Xi\left(t\right)\right)\right)}=\lim_{t\to 0^{+}}\frac{1}{\sqrt{2\pi}}\frac{t^{\delta}e^{-t/2}}{\phi\left(\Phi^{-1}\left(1-\Xi\left(t\right)\right)\right)}\\
			\stackrel{\text{L'Hospital's rule}}{=}&\lim_{t\to 0^{+}}\frac{t^{\delta}e^{-t/2}}{\sqrt{2\pi}}\frac{\delta t^{-1}-1/2}{\Phi^{-1}\left(1-\Xi\left(t\right)\right)\xi\left(t\right)}=\lim_{t\to 0^{+}}\frac{\left(\delta t^{-1}-1/2\right)t^{\delta+1/2}}{\Phi^{-1}\left(1-\Xi\left(t\right)\right)}\\
			=&\lim_{t\to 0^{+}}\frac{\delta t^{\delta-1/2}}{\Phi^{-1}\left(1-\Xi\left(t\right)\right)}.
		\end{align*}
		To deal with the denominator, we use Chebyshev's approximation for the inverse of the Gaussian CDF \citep{blair1976rational}. Specifically, $\Phi^{-1}\left(1-\Xi\left(t\right)\right)=\sqrt{-2\log \Xi\left(t\right)}(1+o(1))$ for $\Xi\left(t\right)\to0$. Thus we have
		\begin{align*}
			\lim_{t\to 0^{+}}\frac{\delta t^{\delta-1/2}}{\Phi^{-1}\left(1-\Xi\left(t\right)\right)}&=\lim_{t\to 0^{+}}\frac{\delta t^{\delta-1/2}}{\sqrt{-2\log\Xi\left(t\right)}}.
		\end{align*}
		We approximate the CDF function $\Xi\left(t\right)$ by the limiting of $\Xi\left(t\right)$ as $t\to 0^{+}$. The CDF function $\Xi\left(t\right)=\gamma\left(1/2,t/2\right)/\Gamma\left(1/2\right)$, where $\gamma\left(\cdot,\cdot\right)$ is the \textit{incomplete gamma function} and $\Gamma\left(\cdot\right)$ is the \textit{gamma function}. Using the approximation of the incomplete gamma function, we have $\gamma\left(1/2,t/2\right)=\sqrt{2t}(1+o(1))$ as $t\to 0^{+}$, and
		\begin{equation*}
			\lim_{t\to 0^{+}}\frac{\delta t^{\delta-1/2}}{\sqrt{-2\log\Xi\left(t\right)}}=\lim_{t\to 0^{+}}\frac{\delta t^{\delta-1/2}}{\sqrt{-\log t}}=0.
		\end{equation*}
		Now we are ready to prove the statement on sensitivity. Because the function $\frac{\xi\left(t\right)}{\phi\left(\Phi^{-1}\left(1-\Xi\left(t\right)\right)\right)}$ is continuous and converges to $0$ as $t\to \infty$, there exists a threshold $C_3>0$ and the function $\frac{\xi\left(t\right)}{\phi\left(\Phi^{-1}\left(1-\Xi\left(t\right)\right)\right)}$ has an upper bound $C_1>0$ on $\left[C_3,\infty\right)$. By the fact that the function $t^{1/2+\delta}\frac{\xi\left(t\right)}{\phi\left(\Phi^{-1}\left(1-\Xi\left(t\right)\right)\right)}$ is continuous and converges to $0$ as $t\to 0$. For the enough small threshold $C_3>0$, the function $t^{1/2+\delta}\frac{\xi\left(t\right)}{\phi\left(\Phi^{-1}\left(1-\Xi\left(t\right)\right)\right)}$ has an upper bound $C_2>0$ on $\left(0,C_3\right]$. Thus, the function $\frac{\xi\left(t\right)}{\phi\left(\Phi^{-1}\left(1-\Xi\left(t\right)\right)\right)}$ is bounded by $C_2 t^{-1/2-\delta}$ on $\left(0,C_3\right]$. Then, we have inequality,
		\begin{align*}
			&\left|\Phi^{-1}\{1-\Xi\left[T\left(\mathcal{S}\right)\right]\}-\Phi^{-1}\{1-\Xi\left[T\left(\mathcal{S}^{\prime}\right)\right]\}\right|\\
			\leq &\left|\int_{T\left(\mathcal{S}^{\prime}\right)}^{T\left(\mathcal{S}\right)}\frac{\xi\left(t\right)}{\phi\left(\Phi^{-1}\left(1-\Xi\left(t\right)\right)\right)}dt\right|\\
			\leq&\left|\int_{\max\{T\left(\mathcal{S}^{\prime}\right),C_3\}}^{\max\{T\left(\mathcal{S}\right),C_3\}}\frac{\xi\left(t\right)}{\phi\left(\Phi^{-1}\left(1-\Xi\left(t\right)\right)\right)}dt\right|+\left|\int_{\min\{T\left(\mathcal{S}^{\prime}\right),C_3\}}^{\min\{T\left(\mathcal{S}\right),C_3\}}\frac{\xi\left(t\right)}{\phi\left(\Phi^{-1}\left(1-\Xi\left(t\right)\right)\right)}dt\right|\\
			\leq&\left|\int_{\max\{T\left(\mathcal{S}^{\prime}\right),C_3\}}^{\max\{T\left(\mathcal{S}\right),C_3\}}C_1dt\right|+\left|\int_{\min\{T\left(\mathcal{S}^{\prime}\right),C_3\}}^{\min\{T\left(\mathcal{S}\right),C_3\}}C_2t^{-1/2-\delta}dt\right|\\
			\leq&C_1\|\max\{T\left(\mathcal{S}^{\prime}\right),C_3\}-\max\{T\left(\mathcal{S}\right),C_3\}\|+C_2\left|\int_{\min\{T\left(\mathcal{S}^{\prime}\right),T\left(\mathcal{S}\right),C_3\}}^{\min\{T\left(\mathcal{S}^{\prime}\right),T\left(\mathcal{S}\right),C_3\}+\|T\left(\mathcal{S}\right)-T\left(\mathcal{S}^{\prime}\right)\|}t^{-1/2-\delta}dt\right|\\
			\leq&M^2/n C_1+C_2\left|\int_{0}^{\|T\left(\mathcal{S}\right)-T\left(\mathcal{S}^{\prime}\right)\|}t^{-1/2-\delta}dt\right|\\
			\leq&M^2/n C_1+\frac{C_2}{1/2-\delta}\left(M^2/n\right)^{1/2-\delta},
		\end{align*}
		where in the fourth inequality, we use the fact that $t^{-1/2-\delta}$ is decreasing, and in the fifth inequality, we use the fact that $\sup\|T\left(\mathcal{S}^{\prime}\right)-T\left(\mathcal{S}\right)\|\leq M^2/n$.
	\end{proof}
	
	\section{Proof of Lemma \ref{report_nosiy_min_al_lemma}}
	\begin{proof}
		We first show that \textit{report noisy min algorithm} is $\left(\epsilon,\delta\left(\epsilon\right)\right)$-DP, where $\epsilon>0$ and $$\delta\left(\epsilon\right)=\Phi\left(-\frac{\epsilon}{\mu}+\frac{\mu}{2}\right)-e^{\epsilon}\Phi\left(-\frac{\epsilon}{\mu}-\frac{\mu}{2}\right).$$
		Using the idea in \cite{dwork2014algorithmic}, we first fix any $i\in\{1,\dots,n\}$. Let $f_1=G^{-1}\left(p_1\right),\dots,f_n=G^{-1}\left(p_1\right)$ to denote the functions when the data is $\mathcal{S}$ and $f_1^{\prime}=G^{-1}\left(p_1^{\prime}\right),\dots,f_n^{\prime}=G^{-1}\left(p_n^{\prime}\right)$ to denote the functions when the data is $\mathcal{S}^{\prime}$, where $\mathcal{S}^{\prime}$ is a neighborhood of $\mathcal{S}$. By the fact that the $G$ function is monotone increasing, it is enough to consider reporting the noisy minimum of $\{f_1,\dots,f_n\}$.
		
		Fix the random noise $Z_i$, $Z_1,\dots,Z_{i-1},Z_{i+1},\dots,Z_n$, which are draw independently from $N\left(0,8\Delta^2/\mu^2\right)$. Define 
		$$Z^{*}=\max_{Z_i}:f_i+Z_i\leq f_j+Z_j\text{ }\forall j\neq i. $$ Then, $i$ will be the algorithm's output when the data is $\mathcal{S}$ if and only if $Z_i\leq Z^{*}$.
		
		We have for $j\neq i$,
		\begin{align*}
			f_i+Z^{*}&\leq f_j+Z_j\\
			\Rightarrow -\Delta+f_i^{\prime}+Z^{*}\leq f_i+Z^{*}&\leq f_j+Z_j\leq f_j^{\prime}+Z_j+\Delta\\
			\Rightarrow f_i^{\prime}+Z^{*}-2\Delta & \leq f_j^{\prime}+Z_j.
		\end{align*}
		Then, if $Z_i\leq Z^{*}-2\Delta$, $i$ will be the algorithm's output when the data is $\mathcal{S}^{\prime}$. 
		\begin{claim}
			$1-\Phi\left(x+\mu\right)\geq e^{-\epsilon}\left[1-\Phi\left(x\right)-\delta\left(\epsilon\right)\right]$, where $\epsilon>0$ and $\delta\left(\epsilon\right)=\Phi\left(-\epsilon/\mu+\mu/2\right)-e^{\epsilon}\Phi\left(-\epsilon/\mu-\mu/2\right)$.
		\end{claim}
		By the claim and letting $x=-Z^{*}\mu/\left(2^{3/2}\Delta\right)$, we have the following relation,
		\begin{align*}
			\mathbf{P}\left(Z_i\geq Z^{*}+2\Delta\right)&\geq e^{-\epsilon}\left[\mathbf{P}\left(Z_i\geq Z^{*}\right)-\delta\left(\epsilon\right)\right]\\
			\Rightarrow \mathbf{P}\left(i\mid\mathcal{S}^{\prime},\boldsymbol{Z}_{-i}\right)\geq\mathbf{P}\left(Z_i\geq Z^{*}+2\Delta\right)&\geq e^{-\epsilon}\left[\mathbf{P}\left(Z_i\geq Z^{*}\right)-\delta\left(\epsilon\right)\right]\geq e^{-\epsilon}\left[\mathbf{P}\left(i\mid\mathcal{S},\boldsymbol{Z}_{-i}\right)-\delta\left(\epsilon\right)\right],
		\end{align*}
		where we use $\mathbf{P}\left(i\mid\mathcal{S},\boldsymbol{Z}_{-i}\right)$ and $\mathbf{P}\left(i\mid\mathcal{S}^{\prime},\boldsymbol{Z}_{-i}\right)$ to denote the probabilities of output $i$ using data $\mathcal{S}$ and $\mathcal{S}^{\prime}$, respectively. We use $\boldsymbol{Z}_{-i}$ to denote $\{Z_1,\dots,Z_{i-1},Z_{i},\dots,Z_{m}\}$. Then, after taking the expectation of $\boldsymbol{Z}_{-i}$, we conclude,
		$$P\left(i\mid\mathcal{S}\right)\leq e^{\epsilon}P\left(i\mid\mathcal{S}^{\prime}\right)+\delta\left(\epsilon\right).$$
		Then, we turn to prove the claim. Let $g\left(x\right)=1-\Phi\left(x+\mu\right)- e^{-\epsilon}\left[1-\Phi\left(x\right)-\delta\left(\epsilon\right)\right]$. Take the derivative of $g\left(x\right)$, we have $$\left.\frac{d}{dx}g\left(x\right)\right\vert_{x=\frac{\epsilon}{\mu}-\frac{\mu}{2}}=0,\text{ }g\left(x\right)<0\text{ if }x<\frac{\epsilon}{\mu}-\frac{\mu}{2},\text{ }g\left(x\right)>0\text{ if }x>\frac{\epsilon}{\mu}-\frac{\mu}{2}.$$
		Then we conclude that $g\left(x\right)$ reaches global minimize at $x=\frac{\epsilon}{\mu}-\frac{\mu}{2}$, and $g\left(\frac{\epsilon}{\mu}-\frac{\mu}{2}\right)=0$ by calculus. Then the claim holds. By Corollary 2.13 in \cite{dong2021gaussian}, report noisy min index is $2^{-1/2}\mu$-GDP. Using the composition theorem, we conclude report noisy min algorithm is $\mu$-GDP.
	\end{proof}

\section{Proof of Theorem \ref{thm: dpadapt_ed}}
 \begin{proof}
     The Report Noisy Min algorithm with Laplace noise $\lambda=\Delta\sqrt{10m\log\left(1/\delta\right)}/\epsilon$ is $\left(2\epsilon/\sqrt{10m\log\left(1/\delta\right)},0\right)$-differential private by Lemma 2.4 in \citep{dwork2021differentially}.
     \begin{lemma}[Advanced Composition Theorem by \cite{dwork2014algorithmic}]
     \label{lem:com}
         For all $\epsilon,\delta\geq 0$ and $\delta^{\prime}>0$, running $l$ mechanisms sequentially that are each $\left(\epsilon,\delta\right)$-differentially private preserves $\left(\epsilon\sqrt{2l\log\left(1/\delta^{\prime}\right)}+l\epsilon\left(e^{\epsilon}-1\right),l\delta+\delta^{\prime}\right)$-differential privacy.
     \end{lemma}
     By Lemma \ref{lem:com}, the peeling algorithm is $\left(\tilde{\epsilon},\delta\right)$-differentially private, where
     \begin{align*}
     \tilde{\epsilon}&=\frac{2\epsilon}{\sqrt{10m\log\left(1/\delta\right)}}\sqrt{2m\log\left(1/\delta\right)}+m\frac{2\epsilon}{\sqrt{10m\log\left(1/\delta\right)}}\left(e^{\frac{2\epsilon}{\sqrt{10m\log\left(1/\delta\right)}}}-1\right)\\
     &=\left[\frac{2}{\sqrt{5}}+\frac{2\sqrt{m}}{\sqrt{10\log\left(1/\delta\right)}}\left(e^{\frac{2\epsilon}{\sqrt{10m\log\left(1/\delta\right)}}}-1\right)\right]\epsilon\\
     &\leq\left(\frac{2}{\sqrt{5}}+1.034\frac{2\sqrt{m}}{\sqrt{10\log\left(1/\delta\right)}}\frac{2\epsilon}{\sqrt{10m\log\left(1/\delta\right)}}\right)\epsilon\\
     &=\left(\frac{2}{\sqrt{5}}+1.034\frac{4\epsilon}{10\log\left(1/\delta\right)}\right)\epsilon,
     \end{align*}
     where we use the relation $e^x-1\leq 1.034x$ for $0\leq x\leq 0.0660$ and $2\epsilon/\sqrt{10m\log\left(1/\delta\right)}\leq 0.0660$ for $\epsilon\leq 0.5,\delta\leq 0.1$ and $m\geq 10$. Furthermore,
     \begin{align*}
         \tilde{\epsilon}\leq \left(\frac{2}{\sqrt{5}}+1.034\frac{4\times 0.5}{10\log\left(10\right)}\right)\epsilon\leq\epsilon.
     \end{align*}
     Then by post-processing property of $\left(\epsilon,\delta\right)$-differential privacy, the DP-AdaPT algorithm with Laplace noise is $\left(\epsilon,\delta\right)$-differential privacy.
 \end{proof}
 
	\section{Proof of Theorem \ref{thm:DP-AdaPT}}
	\begin{proof}
		We begin with a technique lemma.  Let $\left[n\right]$ denote the set $\{1,\dots,n\}$.
		\begin{lemma}[\cite{lei2018adapt}]
			Suppose that, conditionally on the $\sigma$-field $\mathcal{G}_{-1}$, $b_1,\dots,b_n$ are independent Bernoulli random variables with $\mathbb{P}\left(b_i=1\mid\mathcal{G}_{-1}\right)=\rho_i \geq\rho>0$, almost surely. Also, suppose that $\left[n\right]\supseteq\mathcal{C}_0\supseteq\mathcal{C}_1\supseteq\dots$, with each subset $\mathcal{C}_{t}$ measurable with respect to
			$$\mathcal{G}_{t}=\sigma\left(\mathcal{G}_{-1},\mathcal{C}_t,\left(b_i\right)_{i\notin\mathcal{C}_t},\sum_{i\in\mathcal{C}_t}b_i\right).$$
			If $\hat{t}$ is an almost-surely finite stopping time with respect to the filtration $\left(\mathcal{G}_{t}\right)_{t\geq 0}$, then
			$$\mathbb{E}\left[\frac{1+\left|\mathcal{C}_{\hat{t}}\right|}{1+\sum_{i\in\mathcal{C}_{\hat{t}}b_i}}\bigm|\mathcal{G}_{-1}\right]\leq\rho^{-1}.$$
		\end{lemma}
		
		Let $\hat{t}$ denote the step at which we stop and reject. Then
		$$\text{FDP}_{\hat{t}}=\frac{V_{\hat{t}}}{R_{\hat{t}}\vee 1}=\frac{1+U_{\hat{t}}}{R_{\hat{t}}\vee 1}\frac{V_{\hat{t}}}{1+U_{\hat{t}}}\leq\frac{1+A_{\hat{t}}}{R_{\hat{t}}\vee 1}\frac{V_{\hat{t}}}{1+U_{\hat{t}}}\leq \alpha\frac{V_{\hat{t}}}{1+U_{\hat{t}}},$$
		where we define 
		$$U_t=\left|\{i_j,j=1,\dots,m:H_{i_j}\text{ is null and }\tilde{p}_{i_j}\geq 1-s_t\left(\boldsymbol{x}_{i_j}\right)\}\right|,$$
		and 
		$$V_t=\left|\{i_j,j=1,\dots,m:H_{i_j}\text{ is null and }\tilde{p}_{i_j}\leq s_t\left(\boldsymbol{x}_{i_j}\right)\}\right|,$$
		and we use the fact $U_{t}\leq A_{t}$ and the stopping criterion $\left(1+A_{\hat{t}}\right)/\left(R_{\hat{t}}\vee 1\right)\leq\alpha$. It is enough to consider $V_{\hat{t}}/\left(1+U_{\hat{t}}\right)$.
		
		We use $\tilde{p}_{i_j}$ to denote the noisy $p$-values that are subject to release, for $j=1,\dots,m$. Let $\tilde{m}_{i_j}=\min\{\tilde{p}_{i_j},1-\tilde{p}_{i_j}\}$, $b_{i_j}=1\{\tilde{p}_{i_j}\geq 0.5\}$ and $\tilde{p}_{i_j}=b_{i_j}\left(1-\tilde{m}_{i_j}\right)+\left(1-b_{i_j}\right)\tilde{m}_{i_j}$. We use $\mathcal{P}$ to denote the indexes returned by the mirror peeling algorithm. Let $\mathcal{C}_t=\{i\in\mathcal{H}_0,i\in\mathcal{P}:\tilde{p}_{i}\notin\left(s_t\left(\boldsymbol{x}_i\right),1-s_t\left(\boldsymbol{x}_i\right)\right)\}$ be the index of the null $p$-values which are unavailable to the analyst. Then, we have the relation $U_{t}=\sum_{i\in\mathcal{C}_t}b_i$ and $V_t=\sum_{i\in\mathcal{C}_t}\left(1-b_i\right)=\left|\mathcal{C}_t\right|-U_t$.

		We then construct the auxiliary filtration $\left(\mathcal{G}_{t}\right)_{t\geq -2}$. First, define the $\sigma$-field,
		$$\mathcal{G}_{-2}=\sigma\left(\{\boldsymbol{x}_i,m_i\}_{i=1}^{n}\right),$$
		where $m_i=\min\{p_i,1-p_i\}$. In the mirror-peeling mechanism, we use $\tilde{p}_{i,j}$ to denote the noisy $p$-value of $\tilde{p}_i$ in the $j$-th peeling mechanism. Define the peeling $\sigma$-field,
		$$\mathcal{G}_{-2}\subseteq\mathcal{G}_{-2+\frac{1}{m+1}}\subseteq\dots\subseteq\mathcal{G}_{-1-\frac{1}{m+1}},$$
		where $\mathcal{G}_{-2+\frac{j}{m+1}}=\sigma\left(\mathcal{G}_{-2+\frac{j-1}{m+1}},\{\tilde{m}_{i,j}\}_{i=1}^{n}\right)$ and $\tilde{m}_{i,j}=\min\{\tilde{p}_{i,j},1-\tilde{p}_{i,j}\}$. It is not difficult to see the $j$-th peeling mechanism is measurable with respect to $\mathcal{G}_{-2+\frac{j}{m+1}}$. Then, we define
		$$\mathcal{G}_{-1}=\sigma\left(\mathcal{G}_{-1-\frac{1}{m+1}},\{\tilde{m}_i\}_{i=1}^{n},\left(b_i\right)_{i\notin\mathcal{H}_0}\right),$$
		and
		$$\mathcal{G}_{t}=\sigma\left(\mathcal{G}_{-1},\mathcal{C}_t,\left(b_i\right)_{i\notin\mathcal{C}_t},\sum_{i\in\mathcal{C}_t}b_i\right).$$
		The assumptions of independence and mirror-conservatism guarantee $\mathbb{P}\left(b_i=1\mid\mathcal{G}_{-1}\right)\geq 0.5$
		almost surely for each $i\in\mathcal{H}_0$. The reasons are
		\begin{align*}
			\mathbb{P}\left(b_i=1\mid\mathcal{G}_{-1}\right)&=\mathbb{P}\left(\tilde{p}_i\geq 0.5\mid\mathcal{G}_{-1}\right)=\mathbb{P}\left(\tilde{p}_i\geq 0.5\mid \tilde{m}_i,m_i,\{m_{i,j}\}_{j=1}^{m}\right)\\
			&=\mathbb{P}\left(\tilde{p}_i\geq 0.5\mid \tilde{m}_i\right),
		\end{align*}
		and that $\tilde{p}_i$ is mirror-conservative by Theorem \ref{lem: noisy_p_value_mc}.
		
		Notice that $\tilde{p}_i\in\mathcal{G}_t$ for $\tilde{p}_i\in\left(s_t\left(\boldsymbol{x}_i\right),1-s_t\left(\boldsymbol{x}_i\right)\right)$, $A_t=U_t +\left|i\notin\mathcal{H}_0,i\in\mathcal{P}:\tilde{p}_i\geq 1-s_t\left(\boldsymbol{x}_i\right)\right|$ and $R_t=V_t +\left|i\notin\mathcal{H}_0,i\in\mathcal{P}:\tilde{p}_i\leq s_t\left(\boldsymbol{x}_i\right)\right|$. Then, we conclude $\mathcal{F}_t\subseteq\mathcal{G}_t$. It follows that $\hat{t}=\min\{t:\widehat{\text{FDP}}_{t}\leq\alpha\}$ is a stopping time with respect to $\mathcal{G}_t.$ Then we have
		$$\mathbb{E}\left[\text{FDP}\mid\mathcal{G}_{-1}\right]\leq\alpha\mathbb{E}\left[\frac{V_{\hat{t}}}{1+U_{\hat{t}}}\mid\mathcal{G}_{-1}\right]=\alpha\mathbb{E}\left[\frac{1+\left|\mathcal{C}_{\hat{t}}\right|}{1+U_{\hat{t}}}-1\mid\mathcal{G}_{-1}\right]\leq\alpha.$$
		Finally, notice that $\mathcal{F}_{-1}\subset\mathcal{G}_{-1}$ and use the tower property of conditional expectation.
	\end{proof}
 \label{app:theorem}
\end{document}